\documentclass[journal]{IEEEtran}
\usepackage{amsmath,amsfonts}
\usepackage{algorithmic}
\usepackage{multirow}
\usepackage{algorithm}
\usepackage{array}
\usepackage{textcomp}
\usepackage{stfloats}
\usepackage{subfigure}
\usepackage{amsmath}
\usepackage{epsfig}
\usepackage{epstopdf}
\usepackage{caption}
\usepackage{url}
\usepackage{verbatim}
\usepackage{graphicx}
\usepackage{cite}
\usepackage{color}
\usepackage{booktabs}
\definecolor{DarkGreen}{rgb}{0.1,0.9,0.1} 
\usepackage[%
    pdftex,%
    colorlinks=true,%
    hyperindex,%
    plainpages=false,%
    pagebackref=true,%
    bookmarksopen,%
    bookmarksnumbered %
      ]{hyperref}

\begin{document}

\title{PreMixer: MLP-Based Pre-training Enhanced MLP-Mixers for Large-scale Traffic Forecasting}

\author{
\IEEEauthorblockN{
		Tongtong Zhang, 
		Zhiyong Cui\IEEEauthorrefmark{1}, 
		Bingzhang Wang, 
		Yilong Ren\IEEEauthorrefmark{1},
		Haiyang Yu,
            Pan Deng,
            and Yinhai Wang~\IEEEmembership{, Fellow, IEEE}
  }
\thanks{This work was also partially supported by the National Natural Science Foundation of China (project number:52202378), the Youth Talent Support Program of Beihang University (project number: YWF-22-L-1239), and the Ministry of Transport of PRC Key Laboratory of Transport Industry of Comprehensive Transportation Theory  (project number: MTF2023002). (Corresponding author: Zhiyong Cui and Yilong Ren.)

Tongtong Zhang, Zhiyong Cui, Yilong Ren, and Haiyang Yu are with State Key Laboratory of Intelligent Transportation Systems, Beihang University, Beijing 100191, China (e-mail: tongtongzhang@buaa.edu.cn, zhiyongc@buaa.edu.cn)

Bingzhang Wang and Yinhai Wang are with Civil and Environmental Engineering, University of Washington, Seattle, WA 98195, USA (e-mail: bzwang@uw.edu)

Deng Pan is with School of Software Engineering, Beihang University, Beijing, 100083, China (e-mail: pandeng@buaa.edu.cn)}
}



\maketitle

\begin{abstract}
In urban computing, precise and swift forecasting of multivariate time series data from traffic networks is crucial. This data incorporates additional spatial contexts such as sensor placements and road network layouts, and exhibits complex temporal patterns that amplify challenges for predictive learning in traffic management, smart mobility demand, and urban planning. Recently, large-scale traffic between cities and urban agglomerations has increasingly become a routine part of daily life. Consequently, there is an increasing need to forecast traffic flow across broader geographic regions and for higher temporal coverage. However, current research encounters limitations because of the inherent inefficiency of model and their unsuitability for large-scale traffic network applications due to model complexity. This paper proposes a novel framework, named PreMixer, designed to bridge this gap. It features a predictive model and a pre-training mechanism, both based on the principles of Multi-Layer Perceptrons (MLP). The PreMixer comprehensively consider temporal dependencies of traffic patterns in different time windows and processes the spatial dynamics as well. Additionally, we integrate spatio-temporal positional encoding to manage spatiotemporal heterogeneity without relying on predefined graphs. Furthermore, our innovative pre-training model uses a simple patch-wise MLP to conduct masked time series modeling, learning from long-term historical data segmented into patches to generate enriched contextual representations. This approach enhances the downstream forecasting model without incurring significant time consumption or computational resource demands owing to improved learning efficiency and data handling flexibility. Our framework achieves comparable state-of-the-art performance while maintaining high computational efficiency, as verified by extensive experiments on large-scale traffic datasets. These results underscore our contributions to the traffic prediction in large-scale areas, providing a scalable and efficient solution that leverages deep insights from traffic spatiotemporal data.
\end{abstract}


\section{Introduction}
\begin{figure}[htbp]
    \centering
    \subfigure[Comparison of model performance and computational costs.]{
        \includegraphics[width=0.5\textwidth]{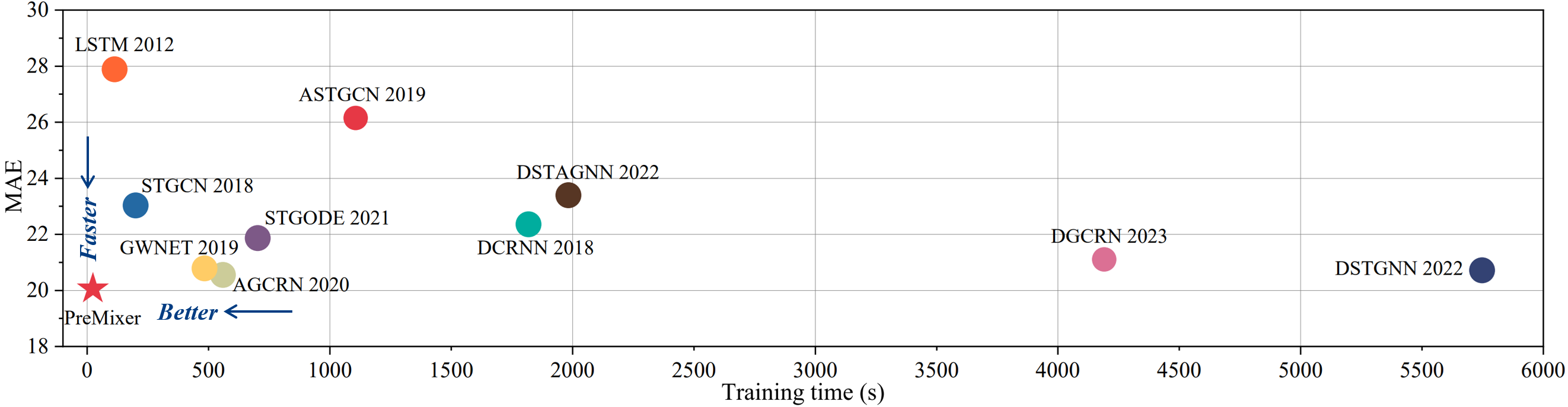}
        \label{fig1a}
    }
    \subfigure[Complex and dynamic long-term spatialtemporal correlations.]{
        \includegraphics[width=0.5\textwidth]{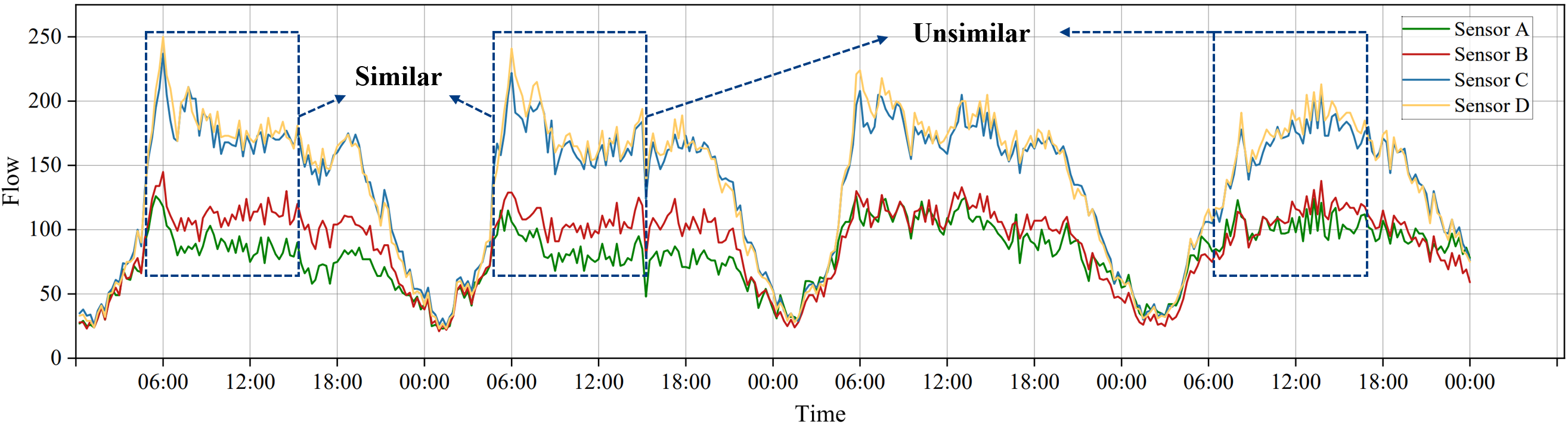}
        \label{fig1b}
    }
    \caption{A comprehensive comparison with recent methods and the long-term traffic pattern in spatialtemporal data. (a) provides a comparative analysis of recent methods, focusing on prediction accuracy and computational cost. (Evaluation is conducted on GLA dataset with more than 3000 sensors.) (b) illustrates the dynamics of periodic changes in traffic patterns, highlighting both similarities and differences.}
    \label{fig1}
\end{figure}

\label{section1}
\IEEEPARstart{T}{RAFFIC} forecasting is a critical task within urban computing applications, encompassing transportation management, smart mobility demand, and general urban planning \cite{zheng2014urban}. Owing to transitions between free flow, recovery, anomalies and congestion in transportation systems, achieving precise and efficient predictions has remained a significant challenge and an active area of research \cite{polson2017deep}. Compared to classical statistical models, machine learning models such as Support Vector Regression \cite{wu2004travel} and K-nearest Neighbor \cite{cai2016spatiotemporal} are regularly utilized in traffic forecasting problem, because they can capture non-linear relationships and deal with high dimensional data. Deep learning models \cite{ma2015long,cui2020graph}, as a branch of machine learning models, demonstrate the promising capability of capturing the dependency in the high dimensional set of explanatory variables and provide excellent performance in short-term traffic forecasting. For modeling temporal dependency in traffic data, traffic prediction models \cite{cui2020stacked, WANG2024102293} adopt deep learning methods such as LSTM \cite{graves2012long}, TCN \cite{bai2018empirical} and Transformer \cite{vaswani2017attention} as building blocks to improve traffic prediction accuracy and robustness. While traffic forecasting shares similarities with multivariate time series forecasting, it gains additional information through the incorporation of spatial contexts such as sensor locations and road networks, which elucidate the dependencies among traffic input signals. Therefore, alongside designing models that capture temporal patterns in traffic data, developing valid and efficient methods for modeling spatial dynamics is also essential in traffic forecasting. 

Recently, Spatio-Temporal Graph Neural Networks (STGNNs) have been introduced and have quickly gained significant attention. These networks combine Graph Neural Networks (GNNs) with diverse methods for temporal learning, effectively identifying the complex characteristics of correlation and heterogeneity in traffic spatiotemporal data. This combination significantly enhances the accuracy of traffic forecasting \cite{jin2023spatio}. By modeling the underlying correlations between traffic sensor signals, GNNs adeptly discern both local and global dependencies, making them ideally suited to traffic forecasting tasks \cite{liu2023we,jiang2022graph}. Moreover, the precision of the graph structure in STGNNs is crucial for minimizing errors in spatiotemporal data predictions. To overcome the limitations of relying on predefined graph such as adjacency matrices, which can be noisy and inaccurate, several STGNNs incorporated spatial learning modules that utilize graph learning and adaptivity techniques \cite{10.5555/3367243.3367303,10032279}. Furthermore, STGNNs integrated advanced learning frameworks \cite{pan2020spatio,ji2023spatio} to deepen spatiotemporal representation, thereby enabling the models to improve predictive performance. Meanwhile, Transformer-based approaches \cite{jiang2023pdformer, xu2020spatial} have been proven effective for traffic forecasting modeling. The advancement of these sophisticated deep learning models have demonstrated encouraging outcomes in predicting traffic patterns for small-scale traffic networks based on short-term historical data \cite{mallick2020graph, cui2020graph, cini2024taming}.

As transportation evolves, people's travel patterns and daily commutes increasingly involves longer distances, covering wider geographical areas. Large-scale traffic mobility between cities and urban agglomerations has increasingly become a routine part of daily activities \cite{clark2016changes}. Consequently, there is an increasing need to forecast traffic flow across broader geographic regions and for higher temporal coverage. Research on large-scale traffic network prediction is crucial but faces significant challenges. On one hand, many existing traffic forecasting models, developed primarily for smaller datasets, are not scalable when applied to larger sensor networks \cite{liu2024largest}. As shown in Figure\ref{fig1}(a), in pursuit of better prediction results, the complexity of the proposed models has progressively increased. For instance, although STGNNs have demonstrated increased accuracy in traffic forecasting, the complexity of their models escalates with the number of nodes and the extent of temporal coverage. This complexity results in substantial time consumption and high computational demands, significantly affecting their scalability and operational efficiency when deployed across extensive rode networks \cite{liu2023we, zhou2020graph, liu2024largest}. Additionally, the computational demand, which increases proportionally with the length of the input time series, along with the enormous quantity of parameters, significantly hinders Transformer-based approaches \cite{jiang2023pdformer, xu2020spatial} from effectively handling large-scale traffic datasets \cite{wang2024spatiotemporal}.

On the other hand, the complex and non-stationary nature of large-scale traffic systems causes observed spatiotemporal traffic data to exhibit intricate temporal patterns over the long-term, including periodicity and trends. As shown in Figure\ref{fig1b}(b), over a longer observation period, changes in traffic patterns often exhibit similar trends at regular intervals. Longer time spans reveal more distinct features of seasonality, such as the impacts of consistent climate conditions or recurring traffic events on seasonal patterns \cite{WANG2024102293}. Utilizing this long-term information can significantly enhance a model’s ability to discern temporal heterogeneity, leading to more reliable and accurate forecasts \cite{10.1145/3534678.3539396}. Additionally, the indistinguishability of samples in temporal dimensions \cite{shao2022spatial} can be alleviated with contextual information derived from long-term historical time steps. For instance, sensor readings might vary across different time windows but exhibit similar historical patterns. However, directly inputting long-term spatiotemporal data into STGNNs, which are trained end-to-end, results in excessively protracted training and inference times, and optimizing the model also becomes a challenge. Drawing inspiration from the representation techniques used in pre-training model within computer vision \cite{he2022masked} and natural language processing \cite{qiu2020pre}, recent studies \cite{10.1145/3534678.3539396, gao2023spatio} have designed pre-training model that efficiently learn the temporal patterns from extensive historical time series and generate segment-level representations. The hidden representations generated by the pre-training model are leveraged to boost the performance of downstream prediction models, such as Graph WaveNet\cite{10.5555/3367243.3367303}. While these pre-training models enhance downstream traffic prediction tasks, they also impose additional time and computational burdens on the prediction models, particularly for transformer-based pre-training models, which are not suitable for deployment on large-scale traffic networks with numerous sensor locations and complex road networks. 

In this study, we focus our approach on simplicity and effectiveness in handling with implementation and deployment in large-scale traffic datasets, achieving performance comparable to more elaborate STGNNs. The architecture based exclusively on Multi-Layer Perceptron (MLP) captures our attention due to its simplicity, efficiency, and state-of-the-art performance in various fields. These models, known as MLP-Mixer models, can mix information from the input data using interleaved MLP across specific dimensions. Notably, MLP-Mixer models has demonstrated tremendous effectiveness in integrating information and has been applied to fields such as computer vision \cite{tolstikhin2021mlp} and natural language processing \cite{fusco2023pnlp}. In time series forecasting, MLP-Mixer preserves the capability of linear models to capture temporal patterns while simultaneously extracting cross-variate information through mixing operations, such as the dependencies among sensor signals \cite{chen2023tsmixer}. In the context of traffic data, several models \cite{yeh2024random, wang2023st, nie2023contextualizing} have explored MLP-Mixer to better capture spatiotemporal relationships, achieving notable results while maintaining efficiency. Although these models outperform many complex alternatives, there remains a lack of exploration concerning concise models and long-term temporal patterns in large-scale traffic datasets. 

Motivated by these methods, we propose an All-Multi-Layer Perceptron mixer forecasting architecture with MLP-based pre-training model, referred to as PreMixer. Concretely, we develop a prediction model based on the MLP-Mixer, designed to efficiently exploit information across both the temporal and spatial dimensions of the input data. This model consists of an input embedding layer, a TemporalMixer module, a SpatialMixer module, and an output prediction layer. Additionally, to enhance the forecasting capabilities of MLP-Mixers for large-scale traffic datasets, we have developed spatio-temporal positional encoding (STPE) to encode spatiotemporal positional information simultaneously, and we use learnable node embedding to capture local components and reflect the static features of each series. Furthermore, we extend the PreMixer with auxiliary long-term information by pre-training method. Specifically, we design an efficient unsupervised pre-training model that generates useful representations for downstream traffic prediction tasks using the masked autoencoding strategy \cite{he2022masked}. This process involves dividing the long-term historical data into patches of equal length, masking out certain patches, and then independently embedding and reconstructing these patches. We employ a simple patch-wise MLP as a reconstructing encoder, named PIEncoder, rather than a transformer-based model, to embed each patch independently. It’s important to note that this patch-independent embedding method allows the PIEncoder to handle inputs flexibly during the forecasting stage, resulting in lower time consumption and reduced computational resource demands compared to transformer-based pre-training models. In summary, the main contributions are the following:

\begin{enumerate}
    \item We design a simple-yet-effective MLP-based framework, PreMixer, to address large-scale traffic forecasting challenges. The innovation in PreMixer stems from auxiliary embeddings and pre-training method to boost the robustness of MLP-Mixer. To the best of our knowledge, this work is the first systematic exploration on incorporating MLP-based pre-training method into MLP-Mixers forecasting model.
    \item We integrate spatio-temporal positional encoding and learnable node embedding to overcome the problem of the indistinguishability of spatiotemporal data without relying on a predefined graph. This configuration allows the PreMixer to gain additional contextual information while achieving a significant reduction in computational complexity.
    \item We develop an MLP-based pre-training model that captures information from extensive historical data spanning weeks. By independently processing time series patches, this pre-training model can handle inputs of flexible lengths in a patch-wise manner. Consequently, it enhances the downstream model without requiring extensive time consumption or high computational resource demands.
    \item The effectiveness of our proposed framework is validated on real-world large-scale traffic flow datasets. Rigorous experiments consistently demonstrate that PreMixer outperforms state-of-the-art methods. Furthermore, a comprehensive efficiency analysis is conducted to validate the efficiency and scalability of PreMixer.
\end{enumerate}

The rest of this paper is arranged as follows: In Section II we present a literature review of related work. In Section III we outline and analyze the problem of short-term forecasting and introduce the overall architecture of PreMixer. In Section IV, We provide a multi-dimensional performance evaluation of the proposed algorithm. Finally, we summarize our work and future directions.

\section{Literature Review}
\label{section2}

\subsection{Spatiotemporal Forecasting}
The performance of traffic forecasting is significantly impacted by the the modeling of relationships that evolve over time between nodes \cite{wang2024spatiotemporal}. STGNNs have emerged as a particularly successful approach in traffic forecasting, combining the spatial understanding capabilities of GNNs with dynamic temporal learning mechanisms, such as Recurrent Neural Networks (RNNs) or Temporal Convolutional Networks (TCNs), to address intricate correlations present in sptiotemporal data \cite{jin2023spatio}. For example, GWNET \cite{10.5555/3367243.3367303} integrates graph convolution and dilated causal convolution to effectively manage temporal sequences. Other significant research efforts, such as DSTAGNN \cite{lan2022dstagnn} and DGCRN \cite{li2023dynamic}, focus on the autonomous learning of spatial dependencies. Additionally, some studies consider self-attention mechanism in extracting the dynamic spatiotemporal correlations in traffic time series data, including ASTGCN \cite{guo2019attention} and PDFormer \cite{jiang2023pdformer}. However, the complex spatiotemporal modeling solutions of the STGNNs, along with transformer-based models, result in significant time consumption and pose deployment challenges when scaled to large-scale transportation networks. Furthermore, STGNNs face challenges in processing long-term historical time series to capture extensive temporal patterns.

More recently, reasearchs have focused on simplified and scalable deep learning techniques to enhance traffic forecasting performance. SimST \cite{liu2023we} mirrors the effectiveness of GNNs by employing spatial context learning methods, which can be used alongside various temporal models to achieve performance comparable to STGNNs. Additionally, models such as STID \cite{shao2022spatial}, ST-MLP \cite{wang2023st}, RPMixer \cite{yeh2024random} and NexuSQN \cite{nie2023contextualizing} explore the effectiveness of MLP-based models for modeling spatiotemporal data. These models employ simple yet effective methods such as the MLP-Mixer, channel independence, and spatiotemporal feature learning to efficiently achieve impressive prediction results. Although most methods integrate temporal pattern and spatial geographic information, they often overlook the long-term temporal characteristics of spatiotemporal data.

\subsection{Pre-training Method}
Due to the proven success of representation learning of pre-training model in computer vision and natural language processing, scholars are currently investigating its prospective applications in spatiotemporal data mining. Recently, contrastive learning (CL) has emerged as a highly effective pretext task, enhancing the learning capabilities of spatiotemporal representations in STGNNs. For instance, STGCL \cite{liu2022contrastive} proposes integrating CL with the learning of spatiotemporal graph representations and leveraging CL's benefits for traffic forecasting. Additionally, SPGCL \cite{li2022mining} iteratively learns from the current context to discriminate reliable positive neighbor nodes through CL, modeling complex spatial dependencies.  

In the realm of multivariate time series, another innovative self-supervised representation learning approach is masked time series modeling. This task involves patchifying and randomly masking out most of the input signals, then training Transformers to embed time series patches for extracting relationships. The learning objective is to reconstruct the masked content based on the unmasked part patches embedding. For instance, by relating masked modeling to manifold learning, SimMTM \cite{dong2024simmtm} proposes to recover masked time points by the weighted aggregation of multiple neighbors outside the manifold. PatchTST \cite{nie2022time} utilizes a channel-independent approach to process univariate time series by segmenting the series into patches and subsequently masking certain patches directly before reconstructing them to preserve their locality. Recently, for traffic spatiotemporal graph data, STEP \cite{10.1145/3534678.3539396} represents the first attempt to employ masked pre-training techniques to address the inability of STGNNs to learn long-term information. STD-MAE \cite{gao2023spatio} introduces a novel spatiotemporal masking strategy during pre-training, which involves masking separately on spatial and temporal axes. These models utilize the pre-training method to capture long-range temporal dependencies and enhance downstream traffic forecasting. However, these transformer-based pre-training architecture require long-term input data at the forecasting stage, which increases time and computing resource demands, making them less suitable for large-scale traffic datasets.

\subsection{Large-scale Traffic Forecasting}
Large-scale spatiotemporal forecasting is crucial in various domains, including weather forecasting, traffic flow prediction, and environmental monitoring. For instance, Corrformer \cite{wu2023interpretable} effectively addresses the complexities of spatiotemporal correlation and computational demands through an innovative multi-correlation mechanism, thereby enhancing collaborative forecasting across global stations. Similarly, in traffic management, predicting large-scale road networks presents analogous challenges, such as more intricate spatiotemporal interactions and extended data durations \cite{WANG2024102293}. Consequently, models proficient with small-scale datasets often struggle to scale to extensive road network datasets, thus requiring a reevaluation of strategies to maintain both effectiveness and efficiency. The verification results from PFNet \cite{wang2023pfnet} reveal that while some state-of-the-art methods like STGODE \cite{fang2021spatial} exhibit high performance, they demonstrate low efficiency in large-scale traffic dataset. Conversely, models such as GWNET \cite{10.5555/3367243.3367303} and GMAN \cite{zheng2020gman} display higher efficiency but at the cost of reduced performance. 

To address efficiency issues, several studies have examined the constraints of traditional neural network models and introduced novel approaches. Notably, SimST \cite{liu2023we} illustrates that GNNs are not essential for spatial modeling in traffic prediction. It recommends the use of simple local proximity modeling as an alternative, which not only addresses the efficiency issues but also yields significant empirical results in traffic forecasting. PFNet \cite{wang2023pfnet} circumvents the computational complexities associated with GCN-based methods by adopting feature labeling techniques from the field of computer vision, thereby achieving a balance between performance and efficiency in large-scale traffic prediction. STFT \cite{WANG2024102293} conducts a comprehensive analysis of large-scale road networks, pinpointing issues stemming from the computational complexity inherent in spatial and temporal modeling, such as considerable escalations in running time, memory consumption, and parameter count. It incorporates the Transformer architecture along with three critical spatiotemporal components, resulting in enhanced performance and substantial efficiency improvements. Recent studies \cite{nie2023contextualizing, qin2023spatio, yeh2024random} have increasingly concentrated on simpler network architectures, such as MLP, to facilitate the prediction of large-scale traffic networks. These modifications to MLP models, integrated with advanced spatiotemporal modeling techniques, aim to strike a balance between effectiveness and efficiency. However, these models often overlook long-term characteristics in traffic data, which limits the exploitation of long-term features and impedes further improvements in predictive performance.

\section{Methodology}

\subsection{Preliminaries}
\subsubsection{Problem Definition}
Given the $T$-step historical spatial-temporal traffic data
$X_{t-(T+1):t}$, the objective of the traffic forecasting is to predict future time series data $X_{t:t+T}$. We address the task by learning a function $F(\cdot)$ to generate the traffic state as follows:

\begin{equation} \label{eq:1}
    F([X_{t-(T+1):t}]) = [X_{t:t+T}] =\mathcal{Y}
\end{equation}
\\
Here, $X_{i}$ represents the data at the $i$-th time step, where $X_{i} \in \mathbb{R}^{N \times C}$ and $\mathcal{Y} \in \mathbb{R}^{T \times N \times C}$. In this context, $N$ denotes the number of sensor-stations (nodes) and $C$ represents the number of features per sensor station.

\subsection{MODEL ARCHITECTURE}

\begin{figure*}[htbp]
    \centering
    \includegraphics[width=1\textwidth]{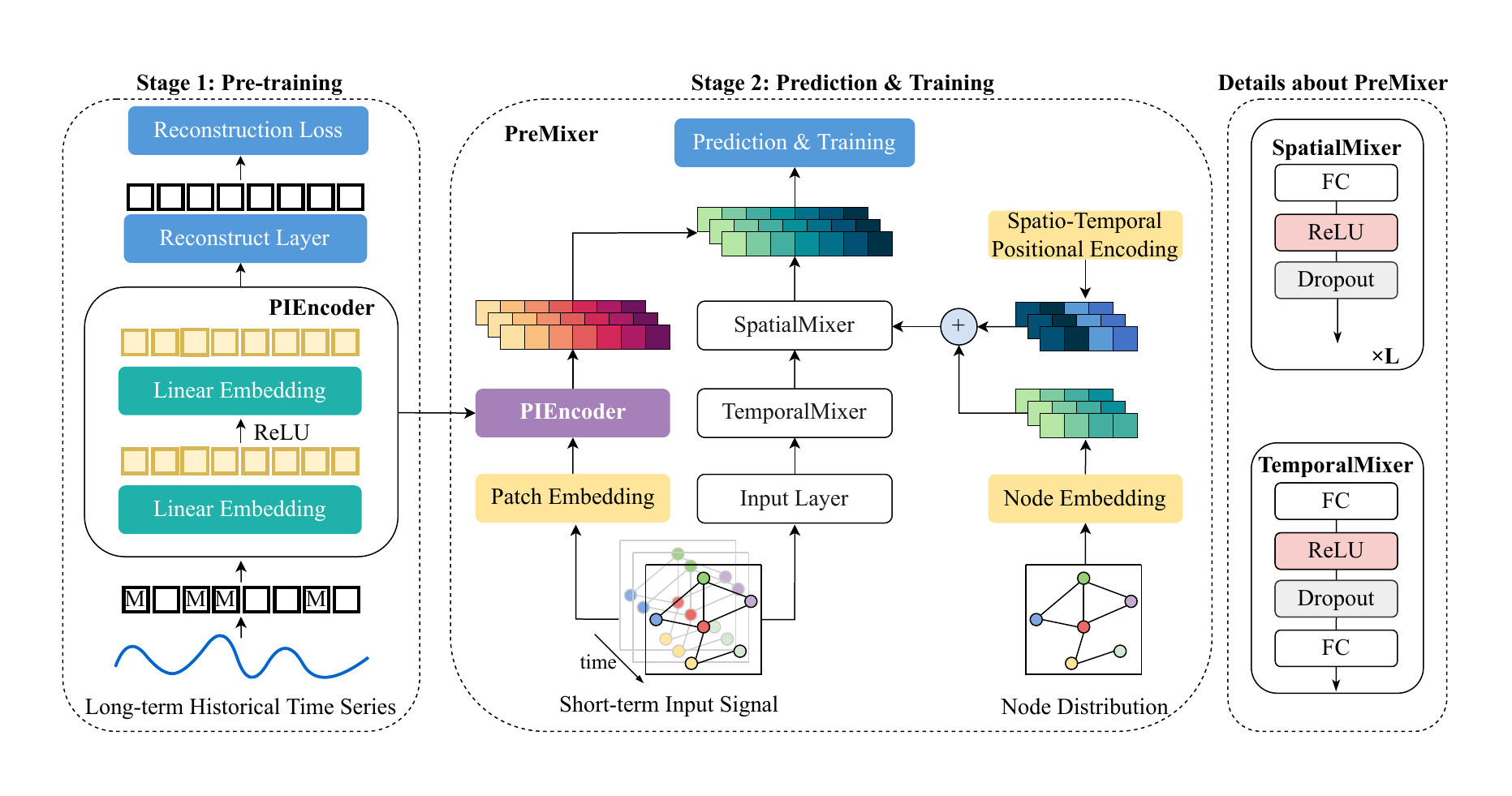}
    \caption{\centering Overall schematic of the PreMixer.}
    \label{fig:model architecture}
\end{figure*}
The structure of PreMixer for predicting traffic state is demonstrated in Figure \ref{fig:model architecture}. The input signal can usually be represented as a spatiotemporal graph, in which the attributes of the graph and the dependencies between sensors evolve over time. The main components of PreMixer include an input layer, TemporalMixer, SpatialMixer, and an output layer. The TemporalMixer models temporal patterns in time series and is shared across nodes. The SpatialMixer is shared by time steps and serves to leverage covariate information between traffic sensor signals. Moreover, we extend the PreMixer with auxiliary information for large-scale traffic forecasting. In addition to the historical observations, we integrate spatio-temporal positional encoding to enhance the model’s recognition capabilities. We also fuse this encoding with learnable node embedding to serve as the pairwise node features, thereby formulating a structured SpatialMixer. Furthermore, we utilize the pre-training model PIEncoder to generate representations with contextual information, enhancing the downstream prediction. The details of these processes are elaborated in the following two subsections.

\subsubsection{The Pre-training Stage}
\begin{figure*}[htbp]
  \includegraphics[width=0.95\textwidth]{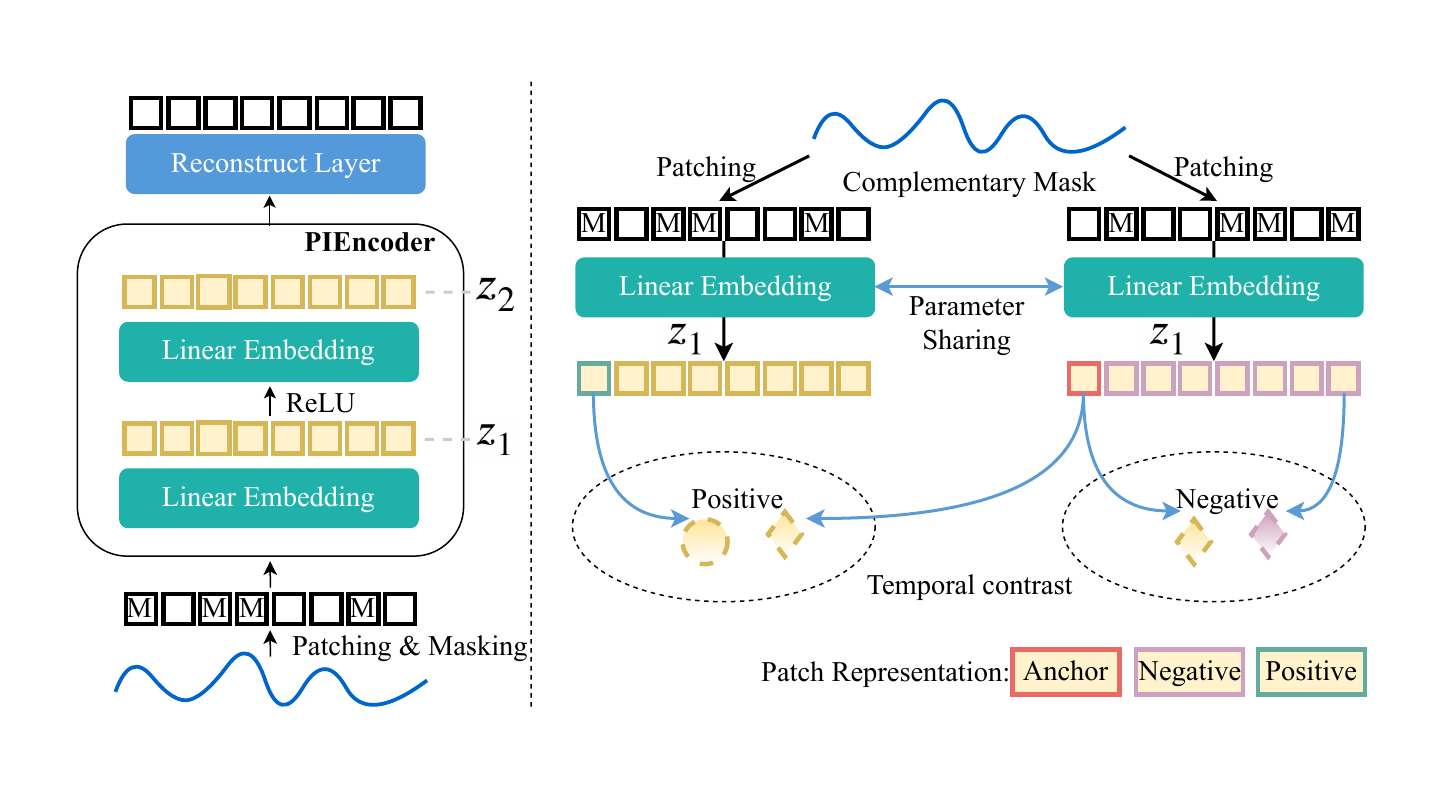}
  \caption{The Pre-training Stage. Left: the overview of the proposed PIEncoder. We segment prolonged time series data spanning the past week into patches and input these into the PIEncoder, which is trained using a masked autoencoding approach. Right: the contrastive learning. we generate two views of the data using a complementary masking strategy and enhance the representations produced by PIEncoder based on temporal contrast.}
  \label{fig:pre-training model}
\end{figure*}

In standard traffic forecasting tasks, the input typically contains short-term historical time series data, and the model struggles to capture long-term temporal patterns, resulting in lower prediction performance compared to models that effectively capture long-term characteristics. Recently, several studies \cite{10.1145/3534678.3539396, gao2023spatio} have explored the use of pre-training models to leverage long-term dependencies and enhance the downstream forecasting model. Similar to these methods, our approach extends the input range to include long-term data $X_{t-\left(T_{\text {long }}-1\right): t} \in \mathbb{R}^{T_{\text {long }} \times N \times C}$ with $T_{\text {long }}$ time steps in the pre-training stage. Furthermore, we segment the original long-term input into non-overlapping $T_{p} = T_{\text {long }}/L$ patches of $L$ time steps. This segmentation strategy enables the pre-training model to capture sufficient information through patch-wise representation learning, thereby increasing efficiency. The tpical pre-training method \cite{10.1145/3534678.3539396, gao2023spatio} embeds the patched input, $\mathcal{X} \in \mathbb{R}^{T_{p} \times N \times L C}$, and then trains Transformers to capture the dependencies between patches by predicting the content of masked patches based on unmasked ones. While the transformer-based pre-training model captures dependencies between patches due to its reliance on self-attention, it may not be the optimal method for time series representation learning and can fail under certain test conditions \cite{lee2023learning}. Additionally, deploying this pre-training model in downstream prediction tasks significantly increases time and computational demands, rendering it unsuitable for large-scale traffic forecasting. To address these issues, we have adapted an MLP-based pre-training model to independently embed the patched input, referred to as PIEncoder. As depicted in Figure \ref{fig:pre-training model} (left), PIEncoder comprises a two-layer MLP with ReLU activation functions and a reconstruction layer based on MLP. This architecture allows every patches to be autoencoded independently, without any inter-patch information exchange.

Specifically, given the patched input $\mathcal{X} \in \mathbb{R}^{T_{p} \times N \times L C}$, we deploy the PIEncoder to embed each patch:

\begin{equation}
\label{eq:2}
    \boldsymbol{z}_{p}^{(i, n)} = \text{PIEncoder}(\boldsymbol{x}_{p}^{(i, n)})
\end{equation}
\\ where $\boldsymbol{x}_{p} = \left\{\boldsymbol{x}_{p}^{(i, n)}\right\}$, $\boldsymbol{z}_{p}=\left\{\boldsymbol{z}_{p}^{(i, n)}\right\}$, and $i=1, \ldots, T_{p}, n=1, \ldots, N$. And $\boldsymbol{x}_{p}^{(i, n)} \in \mathbb{R}^{P}$  and  $\boldsymbol{z}_{p}^{(i, n)} \in \mathbb{R}^{D}$, $P$ denotes the dimension of the input patch, $P = LC$, $D$ denotes the dimension of the output representations. We utilize the channel (node) independence and patch independence method to embed every patches, ensuring that all patches across different nodes share the same model weights and embed independently, i.e, PIEncoder is independent to $i$ and $n$. Unlike typical masked time series models that only embed unmasked patches and predict masked patches using unmasked ones, our approach adopts the patch reconstruction task that autoencodes both masked and unmasked patches.

\textbf{Complementary Contrastive Learning and Objective Function}: CL \cite{nonnenmacher2022utilizing,liu2022contrastive} effectively captures temporal dependencies and dynamic changes in time series data through a self-supervised approach, learning representations by distinguishing between various samples, such as positive and negative ones. This process enhances the model's generalization and discrimination capabilities. To facilitate the integration of adjacent time series information, we utilize complementary CL to enhance patch-wise time series representation. As depicted in Figure \ref{fig:pre-training model} (right), for the patched input $\boldsymbol{x}_{p}$, we apply a complementary masking strategy to generate positive pairs with different views. We set the mask ratio $\boldsymbol{m}$ to 50\%, and treat $\boldsymbol{m} \odot \boldsymbol{x}_{p}$ and $\boldsymbol{(1-m)} \odot \boldsymbol{x}_{p}$ as two views, where $\odot$ denotes element-wise multiplication. This approach allows us to generate two views for CL without designing additional data augmentation. Importantly, the complementary CL does not affect the independent patch representation learning.

We let ${z}_{1}$ and ${z}_{2}$ denote the first and second layer patch-wise representation obtained from the PIEncoder with a two-layer MLP, respectively. We deploy CL at the first layer of the PIEncoder and feed ${z}_{2}$ into a patch-wise linear projection head on the top of the second layer to get a reconstructed result: $\widehat{\boldsymbol{x}}_{p}^{(i, n)}=W \boldsymbol{z}^{(i, n)}$ where  $W \in \mathbb{R}^{P \times D}$.  Then, the reconstruction loss can be written as:

\begin{align}
\label{eq:3}
    \mathcal{L}_{\text{recon}} &= \sum_{i=1}^{T_{p}} \sum_{n=1}^{N}\left\|\boldsymbol{m}^{(i, n)} \odot\left(\boldsymbol{x}_{p}^{(i, n)}-\widehat{\boldsymbol{x}}_{p}^{(i, n)}\right)\right\|_{2}^{2} \nonumber \\
    &+\left\|\left(1-\boldsymbol{m}^{(i, n)}\right) \odot\left(\boldsymbol{x}_{p}^{(i, n)}-\widehat{\boldsymbol{x}}_{p}^{(i, n)}\right)\right\|_{2}^{2} \nonumber \\
    &= \sum_{i=1}^{T_{p}} \sum_{n=1}^{N}\left\|\boldsymbol{x}_{p}^{(i, n)}-\widehat{\boldsymbol{x}}_{p}^{(i, n)}\right\|_{2}^{2}
\end{align}
\\

For CL loss, we define a softmax probability for assessing similarity among all compared pairs in temporal contrastive loss \cite{lee2020mix}. Assume $\boldsymbol{z}_{1}^{(i, n)}=\boldsymbol{z}_{1}^{(i+T_{p}, n)}$  and $\boldsymbol{z}_{1}^{(i+2 T_{p}, n)}$ be the two views of the patch embedding $\boldsymbol{x}^{(i, n)}$. The softmax probability for a pair of patch indices $\left(i, i^{\prime}\right)$ is thus defined:
\begin{equation}
\label{eq:4}
    p\left(\left(i, i^{\prime}\right), n\right)=\frac{\exp \left(\boldsymbol{z}_{1}^{(i, n)} \circ \boldsymbol{z}_{1}^{\left(i^{\prime}, n\right)}\right)}{\sum_{s=1, s \neq i}^{2 T_{p}} \exp \left(\boldsymbol{z}_{1}^{(i, n)} \circ \boldsymbol{z}_{1}^{(s, n)}\right)}
\end{equation}
\\
where we use the dot product $\circ$ as the similarity measure. Consequently, the total contrastive loss can be expressed as follows:

\begin{equation} \label{eq:5}
    \mathcal{L}_{\mathrm{CL}}=\frac{1}{2 T_{p} N} \sum_{i=1}^{2 T_{p}}  \sum_{n=1}^{N}-\log p((i, i+T_{p}), n)
\end{equation}
\\
The final loss is the sum of the reconstruction loss and contrastive loss:

\begin{equation} \label{eq:6}
    \mathcal{L}=\mathcal{L}_{\text {recon }}+\mathcal{L}_{\mathrm{CL}}
\end{equation}

\subsubsection{The Forecasting Stage}
By embedding long-term time series data, the pre-training model generates representations that capture traffic patterns in input signals, thereby enhancing the downstream prediction model through feature fusion. However, the transformer-based pre-training model requires feeding long-term time series data during the forecasting stage, which incurs significant additional time and computational overhead. In contrast, our proposed MLP-based pre-training model can process short-term time series data during the forecasting stage by independently reconstructing each patch representation. This approach optimizes efficiency, as all patches share the same model weights and are embedded independently. The MLP-based pre-training model is applicable to all types of downstream forecasting models. Although STGNNs effectively capture spatial-temporal dependencies and demonstrate strong performance in traffic forecasting, they struggle to scale on large-scale traffic networks with numerous nodes and high temporal coverage. Inspired by the success of the simple yet effective MLP-Mixer in computer vision and time series analysis, we designe a MLP-based forecasting model tailored for large-scale traffic networks.

\textbf{Spatio-Temporal Positional Encoding and Learnable Node Embedding} : Performing mixing operations along temporal and spatial dimensions, the MLP-Mixer model efficiently extracts information and demonstrates superior performance compared to state-of-the-art alternatives \cite{chen2023tsmixer}. However, the challenge with traffic multivariate time series data lies in the numerous nodes and complex temporal patterns, which may hinder the model’s ability to distinguish between various contextual trends. This complexity can lead to overfitting and susceptibility to local suboptimal solutions. For instance, time series data derived from 
traffic flow sensors may vary across different time windows but exhibit similar historical patterns. To address this issue, we employ sinusoidal positional encoding to implement two-dimensional positional encoding \cite{wang2021translating}, which provides essential spatiotemporal positional information to enhance the model’s recognition capabilities.  Given the input $X_{t-(T+1):t} \in \mathbb{R}^{T \times N \times C}
$, we calculate the positional encodings $\mathbf{U}_{\text {pos }} \in \mathbb{R}^{T \times N \times C}$ along the C dimension as follows:

\begin{equation}\label{eq:7}
\left\{\begin{array}{l}
    U_{\text {pos }}(t, n, 2 i)=\sin \left(t / 10000^{4 i / C}\right) \\
    U_{\text {pos }}(t, n, 2 i+1)=\cos \left(t / 10000^{4 i / C}\right) \\
    U_{\text {pos }}(t, n, 2 j+D / 2)=\sin \left(n / 10000^{4 j / C}\right)\\
    U_{\text {pos }}(t, n, 2 j+1+D / 2)=\cos \left(n / 10000^{4 j / C}\right)
\end{array}\right.
\end{equation}
\\
where $t$ and $n$ are the temporal and spatial indices for $T$ and $N$, and $i$,$j$ is an integer in [0,  C /4]. From Equation \ref{eq:7}, it is evident that the first half of the dimensions encodes the temporal positions, while the second half encodes the spatial positions. This position encoding technique offers the advantage of not introducing additional trainable parameters to the neural network, effectively incorporating spatiotemporal context information. Additionally, the adoption of learnable node embedding as positional encodings marks a significant advancement toward more dynamic and context-aware graph modeling \cite{cini2024taming}. We randomly initialize dictionary $\mathbf{E} \in \mathbb{R}^{N \times d_{\text{emb}}}$ as the learnable node embedding, which represents the static features of each series, thereby facilitating a deeper interpretation of spatial information in complex graph structures. We regard spatiotemporal positional information and learnable node embedding as additional sources of information to enhance the forecasting model through various modules via feature fusion or concatenation. Detailed explanations of these integrations are provided in the subsequent subsection.

\textbf{Downstream Spatio-Temporal Predictor} : In this subsection, we present an overview of PreMixer. Initially, we concatenate the short-term input $X_{t-(T+1):t}$ with $\mathbf{U}_{\text {pos }}$, and embed it into a high-dimensional hidden representation using a single MLP. This hidden representation is then processed through two MLP-Mixer based modules, named TemporalMixer and SpatialMixer, designed to extract the temporal and spatial correlations, respectively. As visualized in Figure \ref{fig:model architecture}, the TemporalMixer and SpatialMixer consist of the following blocks: FC (Fully Connected), ReLU (Rectified Linear Unit), and Dropout. Although their architectures appear similar, they serve different functions from two perspectives of spatiotemporal traffic data. Specifically, the TemporalMixer addresses an inherent issue of temporal data mining, while the SpatialMixer models the spatial dependencies with linear complexity. To further enhance the SpatialMixer, we have integrated spatio-temporal positional encoding and learnable node embedding. Finally, an output layer comprising an MLP and a reshaping layer is employed to predict the future traffic state. Additionally, PreMixer has been augmented with auxiliary representations generated by the pre-training model, PIEncoder, to enrich its predictive capabilities.
 
\textit{Input Layer.} As aforementioned, the input layer utilize an MLP to embed the concatenation of the short-term input and the spatio-temporal positional encoding and we reshape the hidden representation:

\begin{equation}
    \mathbf{H}_{c}=\text{Reshape} \left(\operatorname{MLP}\left({X}_{t-T+1: t} \| \mathbf{U}_{\text {pos }}\right)\right)
\end{equation}
\\
where Reshape($\cdot$): $\mathbb{R}^{T \times N \times D} \rightarrow \mathbb{R}^{N \times D T}$. We fuse the representations generated by PreMixer and the augmented representation $ \mathbf{H}^{(0)} \in \mathbb{R}^{ N \times D} $ can be obtained by combining the representations:

\begin{equation}
\mathbf{H}^{(0)}=\operatorname{MLP}\left(\mathbf{Z}_{2}\right) + \mathbf{H}_{c}
\end{equation}
\\
where $\mathbf{Z}_{2}\in \mathbb{R}^{ N \times D}$ , and we utilize the MLP as a semantic projector to transform $\mathbf{Z}_{2}$.

\textit{TemporalMixer}. We design TemporalMixer to encode the complex temporal relations and patterns in traffic time series. The TemporalMixer apply a two-layer feedforward MLP along the time domain and shared by channels(nodes):

\begin{equation}
\mathbf{H}^{(1)}=\mathbf{H}^{(0)}+\sigma\left(\text {LayerNorm}\left(\mathbf{H}^{(0)}\right)\mathbf{W}_{1}+\mathbf{b}_{1}\right)\mathbf{W}_{2}+\mathbf{b}_{2}
\end{equation}
\\
where $\sigma$ is a GELU activation, $\mathbf{W}_{1}$, $\mathbf{W}_{1}$, $\mathbf{b}_{1}$ and $\mathbf{b}_{2}$ are the learnable MLP parameters. 

\textit{SpatialMixer}. The SpatialMixer is shared across time windows and serves to leverage covariate information related to the spatial dependency of traffic multivariate series. We can formulate the SpatialMixer for space mixing as follows:

\begin{equation}
     {\mathbf{H}}^{(l+1)}=\sigma\left(\mathbf{W}^{(l)}_{\text{channel}}\mathbf{H}^{(l)}+\mathbf{b}\right), l \in\{1, \ldots, L\}
\end{equation}
\\
where $\mathbf{W}_{\text{channel}} \in \mathbb{R}^{N^{\prime} \times N}$ is channel mixing parameter. We can see that the SpatialMixer model the multivariate correlations of series without consider temporal information and the absolute position in the sequence, which cause a window-wise multivaluedness. To enable the SptialMixer to become time-varying and adaptive to the interactive temporal context within time windows, we consider a pattern-aware mixing in NexuSQN \cite{nie2023contextualizing} and formulate a structured SptialMixer:

{\small
\begin{align}
\mathbf{m}_{i \leftrightarrow j}^{(l+1)} &= \operatorname{Contx}_{l}\left(\mathbf{h}_{i}^{(l)}, \mathbf{h}_{j}^{(l)}\right) = \Psi\left(\mathbf{h}_{i}^{(l)} \| \mathbf{h}_{j}^{(l)}\right)\left[\mathbf{h}_{i}^{(l)} \| \mathbf{h}_{j}^{(l)}\right], \nonumber \\
\mathbf{m}_{i}^{(l+1)} &= \operatorname{Mix}\left(\mathbf{m}_{i \leftrightarrow j}^{(l+1)} ; \forall j = \{1, \ldots, N\}\right) = \sum_{j=1}^{N} \mathbf{m}_{i \leftrightarrow j}^{(l+1)}, 
\\
\mathbf{h}_{i}^{(l+1)} &= \operatorname{Update}_{l}\left(\mathbf{h}_{i}^{(l)}, \mathbf{m}_{i}^{(l+1)}\right) = \sigma\left(\Theta \mathbf{h}_{i}^{(l)} + \mathbf{m}_{i}^{(l+1)}\right),
\nonumber
\end{align}
}
\\
where $\mathbf{h}_{i}^{(l)}$ is the representation of node $i$ in layer $l$, and $\Theta$ is the feedforward weight. $\Psi$ is the temporal contextualization function and require the pairwise
correlating of node features. The NexuSQN fuse sinusoidal positional encoding and learnable node embedding to get the final spatiotemporal node embedding(STNE) and use the STNE as a query for all available references to contextualize temporal patterns. Note that we denote the fusion of our propose STPE and learnable node embedding as the the pairwise node features. 

Finally, we make predictions using the output latent hidden representations ${\mathbf{H}}^{(l+1)}$ through an MLP regression layer: $\hat{\mathcal{Y}} \in \mathbb{R}^{T \times N \times C}$. Given the ground truth $\mathcal{Y} \in \mathbb{R}^{T \times N \times C}$, we employ mean absolute error as the regression loss:
\begin{align}
\label{equ:13}
     \mathcal{L}_{\text {regression }}&=\mathcal{L}(\hat{\mathcal{Y}}, \mathcal{Y})\nonumber\\
     &=\frac{1}{T N C} \sum_{j=1}^{T} \sum_{i=1}^{N} \sum_{k=1}^{C}\left|\hat{\mathcal{Y}}_{i j k}-\mathcal{Y}_{i j k}\right|
\end{align}
where $N$ is the number of nodes, $T$ is the number of forecasting steps, and $C$ is the dimensionality of the output. Importantly, the parameters of pre-training model PIEncoder is set to a fixed mode during the forecasting phase to lessen computational and memory demands.

\section{Experiments}
\label{section5}
In this section, we perform comprehensive experiments on four real-world large-scale datasets to answer the followering research questions:
\begin{itemize}
  \item \textbf{RQ1}: Can PreMixer outperform state-of-the-art spatialtemporal forecasting models?
  \item \textbf{RQ2}: How effectively does the transfer capability of the pre-trained model in PreMixer perform? 
  \item \textbf{RQ3}: How much imporvement does each component in PreMixer?
  \item \textbf{RQ4}: How does the efficiency of PreMixer compare to the baselines?
  \item \textbf{RQ5}: How does PreMixer perform in real cases?

\end{itemize}

\subsection{Experimental Setup}

\subsubsection{Evaluation Dataset}
\begin{figure*}[htbp]
    \centering
    \includegraphics[width=1\textwidth]{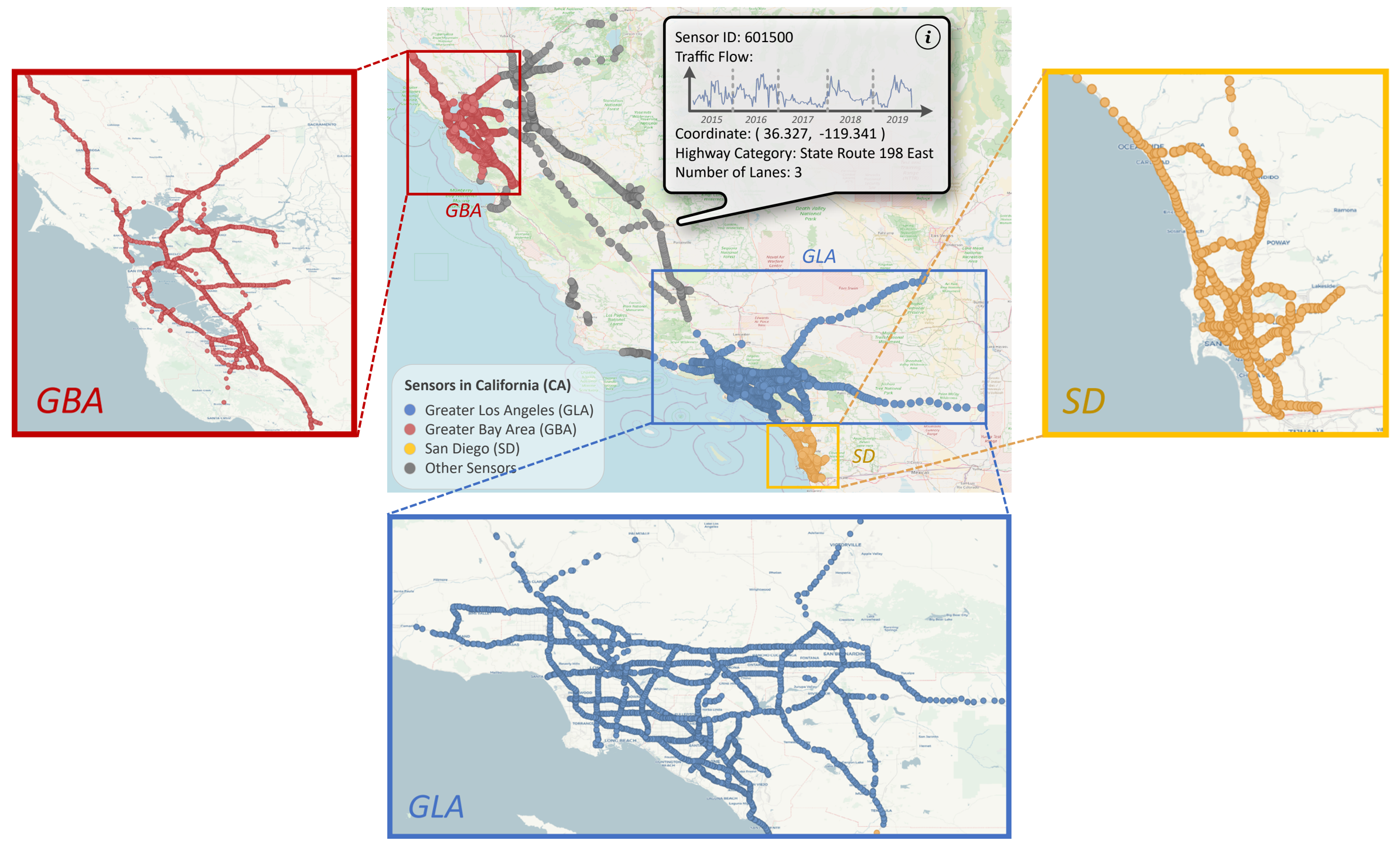}
    \caption{An illustration of the LargeST benchmark dataset \cite{liu2024largest}.}
  \label{fig:compartive}
\end{figure*}

 The training and validation of the PreMixer are performed on the LargeST dataset \cite{liu2024largest}, collected from the California Department of Transportation Performance Measurement System (PeMS) \cite{chen2001freeway}. This dataset includes traffic flow information from 8,600 sensors across California reffered as CA dataset, spanning five years from 2017 to 2021, with data recorded at 5-minute intervals. To facilitate a more comprehensive study on the evolving patterns of transportation systems across various urban agglomerations in California, the dataset is divided into three real-world traffic flow sub-datasets: GLA, GBA, and SD. The GLA encompasses 3,834 sensors in the Greater Los Angeles area, covering five counties: Los Angeles, Orange, Riverside, San Bernardino, and Ventura. The GBA sub-dataset includes 2,352 sensors across 11 counties in the Greater Bay Area: Alameda, Contra Costa, Marin, Napa, San Benito, San Francisco, San Mateo, Santa Clara, Santa Cruz, Solano, and Sonoma. The smallest, SD, contains 716 sensors solely in San Diego County. The LargeST dataset is distinguished not only by its extensive rode network, covering 8,600 sensors, but also by its essential long-term time span and affluent senors data information, with each sensor providing five years of comprehensive metadata. The statistics for the SD, GBA, GLA, and CA datasets are summarized in Table \ref{tab:data_statistics}. We can also observe the geographical distribution of the LargeST benchmark dataset as illustrated in Figure \ref{fig:compartive}.

\begin{table}[htbp]
\centering
\caption{Dataset statistics.}
\label{tab:data_statistics}
\begin{tabular}{lccc}
\hline
Data & \# of nodes & \# of time steps & time range \\
\hline
SD   & 716         & 35,040           & [1/1/2019, 1/1/2020) \\
GBA  & 2,352       & 35,040           & [1/1/2019, 1/1/2020) \\
GLA  & 3,834       & 35,040           & [1/1/2019, 1/1/2020) \\
CA   & 8,600       & 35,040           & [1/1/2019, 1/1/2020) \\
\hline
\end{tabular}
\end{table}

It should be noted that we only adopted 2019 spatiaotemporal data for traffic prediction, adhering to the experimental framework established by \cite{liu2024largest}, to ensure a fair comparison. This data was aggregated from 5-minute to 15-minute intervals, resulting in 96 data points each day. In the experiments, the dataset is split into a training set, a validation set, and a testing set, with a size ratio of 6:2:2. The experiments were executed on a Linux server equipped with an Intel(R) Xeon(R) Gold 6330 CPU and a GeForce GTX 3090 graphics card. During the forecasting phase, we employed Adam optimization with a learning rate of 0.005 and a batch size of 32 to predict the next 12 time steps. Note that in pre-trining stage, each input time series signals in our experiments consisted of 56 patches of 12-length sequences, covering data of the past one week. The hidden dimension of the latent representations in PIEncoder was set to 96. We evaluate the prediction performance of the proposed framework using three common metrics: Mean Absolute Error (MAE), Root Mean Square Error (RMSE), and Mean Absolute Percentage Error (MAPE).

\subsubsection{Baseline} 
To assess both performance and efficiency, we benchmark our proposed model against a selection of baseline models, including 12 of the most representative deep models developed over the years:
\begin{itemize}
  \item HL \cite{liang2021revisiting}: HL is a basic forecasting technique that assumes all future time step predictions are based on the most recent observation.
  \item LSTM \cite{graves2012long}: LSTM is a specialized type of recurrent neural network (RNN) that introduces memory units and gating mechanisms. These features allow it to effectively capture dependencies in long sequences and mitigate the problem of vanishing gradients.
  \item ASTGCN \cite{guo2019attention}: ASTGCN integrates graph convolution, temporal convolution, and an attention mechanism, effectively capturing spatiotemporal dependencies by dynamically adjusting node and timestep weights.
  \item DCRNN \cite{li2017diffusion}: DCRNN integrates graph convolutional networks with recurrent neural networks, capturing spatial dependencies in graph-structured data through diffusion convolution and modeling temporal dependencies with recursive structures.
  \item AGCRN \cite{bai2020adaptive}: AGCRN integrates adaptive graph convolutional networks with gated recurrent units to dynamically learn and adjust node relationships, capture complex spatiotemporal dependencies.
  \item STGCN \cite{10.5555/3304222.3304273}: STGCN builds the model using complete convolutional structures and effectively captures comprehensive spatiotemporal correlations by modeling multi-scale traffic networks.
  \item GWNET \cite{10.5555/3367243.3367303}: GWNET deploys an adaptive graph modelling and is able to handle very long sequences with a stacked dilated 1D convolution component.
  \item STGODE \cite{fang2021spatial}: STGODE proposes a novel continuous representation of GNNs in tensor form for traffic flow forecasting. This approach breaks through the limitations of network depth and enhances the ability to extract longer-range spatiotemporal correlations.
  \item DSTAGNN \cite{lan2022dstagnn}: DSTAGNN designes spatial-temporal aware distance which enhance the representation of the internal dynamic association attributes between nodes of the road network.
  \item DGCRN \cite{li2023dynamic}: DGCRN employs a generative approach to represent the intricate topology of dynamic graphs and adopts a universal training strategy to address the inefficiencies and high resource demands typically associated with RNNs.
  \item \(\mathrm{D}^{2}\mathrm{STGNN}\) \cite{shao2022decoupled}: \(\mathrm{D}^{2}\mathrm{STGNN}\) utilizes a data-driven approach to separate the diffusion and inherent traffic information capturing the dynamic characteristics of road networks.
  \item TSMixer \cite{chen2023tsmixer} TSMixer represents a novel architecture that efficiently extracts information by utilizing mixing operations, constructed using MLPs. This architecture achieves performance comparable to univariate models in long-term time series forecasting benchmarks and real-world large-scale retail demand forecasting tasks.
\end{itemize}

\subsection{Main Results (RQ1)}

\begin{table*}[htbp]
\centering
\caption{Performance comparisons. The highest performance is highlighted in bold, and the second-best is underlined. The missing results of baseline models suggest a memory overflow issue due to the larger number of nodes in the GLA and CA datasets. The data in the 'Average' column represents the mean performance across 12 predicted time steps. Param: the number of learnable parameters. K:kilo $\left(10^{3}\right)$. M: million $\left(10^{6}\right)$.}
\label{tab:prediction result}
\resizebox{\linewidth}{!}{
    \begin{tabular}{cccccc|ccc|ccc|ccc}
        \toprule\multirow{2}{*}{Data} & \multirow{2}*{ Method } & \multirow{2}*{ Param } & \multicolumn{3}{c}{ Horizon 3 } & \multicolumn{3}{c}{ Horizon 6 } & \multicolumn{3}{c}{ Horizon 12 } & \multicolumn{3}{c}{ Average } \\
            \cmidrule(r){4-15}
             & & & MAE & RMSE & MAPE & MAE & RMSE & MAPE & MAE & RMSE & MAPE & MAE & RMSE & MAPE \\
             \midrule \multirow{11}{*}{ SD } &  $\rm{HL}$  & - & 33.61 & 50.97 &  20.77 $\%$  & 57.80 & 84.92 &  37.73 $\%$  & 101.74 & 140.14 &  76.84 $\%$  & 60.79 & 87.40 &  41.88 $\%$  \\
             & LSTM &  98 $\rm{~K}$  & 19.17 & 30.75 &  11.85 $\%$  & 26.11 & 41.28 &  16.53 $\%$  & 38.06 & 59.63 &  25.07 $\%$  & 26.73 & 42.14 &  17.17 $\%$  \\
             & ASTGCN &  2.2 $\rm{M}$  & 20.09 & 32.13 &  13.61 $\%$  & 25.58 & 40.41 &  17.44 $\%$  & 32.86 & 52.05 &  26.00 $\%$  & 25.10 & 39.91 &  18.05 $\%$  \\
             & DCRNN &  373 $\rm{~K}$  & 17.01 & 27.33 &  10.96 $\%$  & 20.80 & 33.03 &  13.72 $\%$  & 26.77 & 42.49 &  18.57 $\%$  & 20.86 & 33.13 &  13.94 $\%$  \\
             & AGCRN &  761 $\rm{~K}$  & 16.05 & 28.78 &  11.74 $\%$  & 18.37 & 32.44 &  13.37 $\%$  & 22.12 & 40.37 &  16.63 $\%$  & 18.43 & 32.97 &  13.51 $\%$  \\
             & STGCN &  508 $\rm{~K}$  & 18.23 & 30.60 &  13.75 $\%$  & 20.34 & 34.42 &  15.10 $\%$  & 23.56 & 41.70 &  17.08 $\%$  & 20.35 & 34.70 &  15.13 $\%$  \\
             & GWNET &  311 $\rm{~K}$  & 15.49 & 25.45 &  \underline{9.90} $\%$  & 18.17 & 30.16 &  11.98 $\%$  & 22.18 & 37.82 &  15.41 $\%$  & 18.12 & 30.21 &  12.08 $\%$  \\
             & STGODE &  729 $\rm{~K}$  & 16.76 & 27.26 &  10.95 $\%$  & 19.79 & 32.91 &  13.18 $\%$  & 23.60 & 41.32 &  16.60 $\%$  & 19.52 & 32.76 &  13.22 $\%$  \\
             & DSTAGNN &  3.9 $\rm{M}$  & 17.83 & 28.60 &  11.08 $\%$  & 21.95 & 35.37 &  14.55 $\%$  & 26.83 & 46.39 &  19.62 $\%$  & 21.52 & 35.67 &  14.52 $\%$  \\
             & DGCRN &  243 $\rm{~K}$  & 15.24 & 25.46 &  10.09 $\%$  & \underline{17.66} & \underline{29.65} &  \textbf{11.77 $\%$}  & \textbf{21.38} & 36.67 &  \underline{14.75} $\%$  & \underline{17.65} & \underline{29.70} &  \textbf{11.89 $\%$}  \\
             &  $\rm{{D}^{2} STGNN}$ &  406 $\rm{~K}$  & \textbf{14.85} & \textbf{24.95} &  9.91 $\%$  & \textbf{17.28} & \textbf{29.05} &  12.17 $\%$  & \underline{21.59} & \textbf{35.55} &  16.88 $\%$  & \textbf{17.38} & \textbf{28.92} &  12.43 $\%$ \\
             & TSMixer &  815 $\rm{~K}$  & 17.13 & 27.42 &  11.35 $\%$  & 19.30 & 31.07 &  12.50 $\%$  & 22.03 & \underline{35.70} &  \textbf{14.26 $\%$}  & 19.06 & 30.66 &  12.55 $\%$  \\
             & PreMixer &  980 $\rm{~K}$  & \underline{15.11} & \underline{25.01} &  \textbf{9.78 $\%$}  & 17.88 & 29.81 &  \underline{11.92} $\%$  & 22.68 & 39.18 &  15.75 $\%$  & 18.02 & 30.22 &  \underline{12.10} $\%$  \\

             \midrule \multirow{11}{*}{GBA} &  $\rm{HL}$  & - & 32.57 & 48.42 &  22.78 $\%$  & 53.79 & 77.08 & 43.01 $\%$  & 92.64 & 126.22 & 92.85$\%$ & 56.44 & 79.82 &  48.87 $\%$  \\
             & LSTM &  98 $\rm{~K}$  & 20.41 & 33.47 &  15.60 $\%$  & 27.50 & 43.64 &  23.25 $\%$  & 38.85 & 60.46 &  37.47 $\%$  & 27.88 & 44.23 &  24.31 $\%$  \\
             & ASTGCN &  22.3 $\rm{M}$  & 21.40 & 33.61 &  17.65 $\%$  & 26.70 & 40.75 &  24.02 $\%$  & 33.64 & 51.21 &  31.15 $\%$  & 26.15 & 40.25 &  23.29 $\%$  \\
             & DCRNN &  373 $\rm{~K}$  & 18.25 & 29.73 &  14.37 $\%$  & 22.25 & 35.04 &  19.82 $\%$  & 28.68 & 44.39 &  28.69 $\%$  & 22.35 & 35.26 &  20.15 $\%$  \\
             & AGCRN &  777 $\rm{~K}$  & 18.11 & 30.19 &  13.64 $\%$  & 20.86 & 34.42 &  16.24 $\%$  & \underline{24.06} & 39.47 &  \underline{19.29} $\%$  & 20.55 & 33.91 &  16.06 $\%$  \\
             & STGCN &  1.3 $\rm{M}$  & 20.62 & 33.81 &  15.84 $\%$  & 23.19 & 37.96 &  18.09 $\%$  & 26.53 & 43.88 &  21.77 $\%$  & 23.03 & 37.82 &  18.20 $\%$  \\
             & GWNET &  344 $\rm{~K}$  & 17.74 & 28.92 &  14.37 $\%$  & 20.98 & 33.50 &  17.77 $\%$  & 25.39 & 40.30 &  22.99 $\%$  & 20.78 & 33.32 &  17.76 $\%$  \\
             & STGODE &  788 $\rm{~K}$  & 18.80 & 30.53 &  15.67 $\%$  & 22.19 & 35.91 &  18.54 $\%$  & 26.27 & 43.07 &  22.71 $\%$  & 21.86 & 35.57 &  18.33 $\%$  \\
             & DSTAGNN &  26.9 $\rm{M}$  & 19.87 & 31.54 &  16.85 $\%$  & 23.89 & 38.11 &  19.53 $\%$  & 28.48 & 44.65 &  24.65 $\%$  & 23.39 & 37.07 &  19.58 $\%$  \\
             & DGCRN &  374 $\rm{~K}$  & 18.09 & 29.27 &  15.32 $\%$  & 21.18 & 33.78 &  18.59 $\%$  & 25.73 & 40.88 &  23.67 $\%$  & 21.10 & 33.76 &  18.58 $\%$  \\
             &  $\rm{{D}^{2} STGNN}$  &  446 $\rm{~K}$  & \textbf{17.20} & \textbf{28.50} &  \textbf{12.22 $\%$}   & 20.80 & 33.53 &  \textbf{15.32 $\%$}  & 25.72 & 40.90 &  19.90 $\%$  & 20.71 & 33.44 &  \underline{15.23} $\%$  \\
             & TSMixer &  3.1 $\rm{M}$  & 17.57 & 29.22 &  14.14 $\%$  & \textbf{19.85} & \textbf{32.64} &  16.95 $\%$  & \textbf{22.27} & \textbf{37.60} &  \textbf{18.63 $\%$}  & \textbf{19.58} & \textbf{32.56} &  \textbf{16.58 $\%$}  \\
             & PreMixer &  1.3 $\rm{~M}$  & \underline{17.35} & \underline{28.88} &  \underline{13.03} $\%$  & \underline{20.21} & \underline{33.15} &  \underline{15.63} $\%$  & 24.26 & \underline{39.20} &  19.87 $\%$  & \underline{20.06} & \underline{32.85} &  15.66 $\%$  \\

             \midrule \multirow{11}{*}{ GLA } &  \rm{HL}  & - & 33.66 & 50.91 &  19.16 $\%$  & 56.88 & 83.54 &  34.85 $\%$  & 98.45 & 137.52 &  71.14 $\%$  & 59.58 & 86.19 &  38.76 $\%$  \\
             & LSTM &  98 $\rm{~K}$  & 20.09 & 32.41 &  11.82 $\%$  & 27.80 & 44.10 &  16.52 $\%$  & 39.61 & 61.57 &  25.63 $\%$  & 28.12 & 44.40 &  17.31 $\%$  \\
             & ASTGCN &  59.1 $\rm{M}$  & 21.11 & 34.04 &  12.29 $\%$  & 28.65 & 44.67 &  17.79 $\%$  & 39.39 & 59.31 &  28.03 $\%$  & 28.44 & 44.13 &  18.62 $\%$  \\
             & DCRNN &  373 $\rm{~K}$  & 18.33 & 29.13 &  10.78 $\%$  & 22.70 & 35.55 &  13.74 $\%$  & 29.45 & 45.88 &  18.87 $\%$  & 22.73 & 35.65 &  13.97 $\%$  \\
             & AGCRN &  792 $\rm{~K}$  & 17.57 & 30.83 &  10.86 $\%$  & \underline{20.79} & 36.09 &  13.11 $\%$  & \underline{25.01} & 44.82 &  \underline{16.11} $\%$  & \underline{20.61} & 36.23 &  \underline{12.99} $\%$  \\
             & STGCN &  2.1 $\rm{M}$  & 19.87 & 34.01 &  12.58 $\%$  & 22.54 & 38.57 &  13.94 $\%$  & 26.48 & 45.61 &  16.92 $\%$  & 22.48 & 38.55 &  14.15 $\%$  \\
             & GWNET &  374 $\rm{~K}$  & \underline{17.30} & \underline{27.72} &  10.69 $\%$  & 21.22 & \underline{33.64} &  13.48 $\%$  & 27.25 & 43.03 &  18.49 $\%$  & 21.23 & 33.68 &  13.72 $\%$  \\
             & STGODE &  841 $\rm{~K}$  & 18.46 & 30.05 &  11.94 $\%$  & 22.24 & 36.68 &  14.67 $\%$  & 27.14 & 45.38 &  19.12 $\%$  & 22.02 & 36.34 &  14.93 $\%$  \\
             & DSTAGNN &  66.3 $\rm{M}$  & 19.35 & 30.55 &  11.33 $\%$  & 24.22 & 38.19 &  15.90 $\%$  & 30.32 & 48.37 &  23.51 $\%$  & 23.87 & 37.88 &  15.36 $\%$  \\
             & DGCRN &  432 $\rm{~K}$  & 17.63 & 28.12 &  \underline{10.50} $\%$  & 21.15 & 33.70 &  \underline{13.06} $\%$  & 26.18 & \underline{42.16} &  17.40 $\%$  & 21.02 & \underline{33.66} &  13.23 $\%$  \\
             &  $\rm{{D}^{2} STGNN}$  &  284 $\rm{~K}$  & 19.31 & 30.07 &  11.82 $\%$  & 22.52 & 35.22 &  14.16 $\%$  & 27.46 & 43.37 &  18.54 $\%$  & 22.35 & 35.11 &  14.37 $\%$  \\
             & TSMixer &  4.6 $\rm{M}$  & 20.38 & 224.82 &  13.62 $\%$  & 22.90 & 229.86 &  15.51 $\%$  & \textbf{23.63} & 135.09 &  \textbf{15.56} $\%$  & 22.12 & 207.68 &  14.87 $\%$  \\
             & PreMixer &  1.8 $\rm{~M}$  & \textbf{16.75} & \textbf{27.25} &  \textbf{9.96 $\%$}  & \textbf{20.30} & \textbf{33.05} &  \textbf{12.25} $\%$  & 25.15 & \textbf{40.99} &  16.20 $\%$  & \textbf{20.19} & \textbf{32.88} &  \textbf{12.39 $\%$}  \\

             \midrule \multirow{6}{*}{ \rm{CA} } &  \rm{HL}  & - & 30.72 & 46.96 &  20.43 $\%$  & 51.56 & 76.48 &  37.22 $\%$  & 89.31 & 125.71 &  76.80 $\%$  & 54.10 & 78.97 &  41.61 $\%$  \\
             & LSTM &  98 $\rm{~K}$  & 19.01 & 31.21 &  13.57 $\%$  & 26.49 & 42.54 &  20.62 $\%$  & 38.41 & 60.42 &  31.03 $\%$  & 26.95 & 43.07 &  21.18 $\%$  \\
             & DCRNN &  373 $\rm{~K}$  & 17.52 & 28.18 &  \underline{12.55} $\%$  & 21.72 & 34.19 &  16.56 $\%$  & 28.45 & 44.23 &  23.57 $\%$  & 21.81 & 34.35 &  16.92 $\%$  \\
             & STGCN &  4.5 $\rm{M}$  & 19.14 & 32.64 &  14.23 $\%$  & 21.65 & 36.94 &  16.09 $\%$  & 24.86 & \underline{42.61} &  19.14 $\%$  & 21.48 & 36.69 &  16.16 $\%$  \\
             & GWNET &  469 $\rm{~K}$  & \underline{16.93} & \underline{27.53} &  13.14 $\%$  & 21.08 & \underline{33.52} &  16.73 $\%$  & 27.37 & 42.65 &  22.50 $\%$  & 21.08 & \underline{33.43} &  16.86 $\%$  \\
            & STGODE &  1.0 $\rm{M}$  & 17.59 & 31.04 &  13.28 $\%$  & 20.92 & 36.65 &  \underline{16.23} $\%$  & 25.34 & 45.10 &  20.56 $\%$  & 20.72 & 36.65 &  16.19 $\%$  \\
            & TSMixer &  9.5 $\rm{M}$  & 18.40 & 106.28 &  14.30 $\%$  & \underline{19.77} & 73.98 &  15.30 $\%$  & \textbf{22.56} & 87.56 &  \underline{17.80} $\%$  & \underline{19.86} & 90.20 &  \underline{15.79} $\%$  \\
            & PreMixer &  3.3 $\rm{~M}$  & \textbf{15.65} &\textbf{25.97}  &  \textbf{11.31 $\%$}  & \textbf{18.53} & \textbf{30.69} &  \textbf{13.57 $\%$}  & \underline{22.59} & \textbf{37.56} &  \textbf{17.26 $\%$}  & \textbf{18.42} & \textbf{30.54} &  \textbf{13.60 $\%$}  \\
        \bottomrule
    \end{tabular}
}
\end{table*}
The prediction results tested of MAE, RMSE, and MAPE on the SD, GBA, GLA, and CA datasets with respect to different predicted horizons are displayed Table \ref{tab:prediction result} as well as the average values. Basic approaches like HL and LSTM, applicable to all sub-datasets with minimal parameter requirements, exhibit the poorest prediction performance due to their failure to account for dependencies among sensor signals. Since incorporating graph structure learning, the AGCRN utilizing RNNs method and the GWNET based on TCNs approach have outperformed other baselines on the SD, GBA, and GLA datasets. We also observe that newer methods like DGCRN and \(\mathrm{D}^{2}\mathrm{STGNN}\) perform well, leveraging the dynamic nature of node relationship in traffic rode networks. However, it is important to highlight that only four spatiotemporal models — DCRNN, GWNET, STGCN, and STGODE — are applicable to the largest dataset, CA. Due to the vast number of nodes in the CA dataset, the parameters of other baseline models increase rapidly, necessitating more computational memory even to conduct them with smaller batch size on our devices. Consequently, these models are not suitable for deployment in large-scale urban road networks and real-world production environments.

In the case of TSMixer, it is comparable with the best methods on the two smaller datasets (SD and GBA). However, it shows significantly poorer performance in terms of RMSE on the larger datasets (GLA and CA), indicating a potential overfitting to MAE. Thanks to its reliance on a linear architecture, TSMixer can scale to larger datasets even with a high parameter count. PreMixer, on the other hand, achieves the best performance on most datasets, demonstrating its excellent capability in large-scale traffic forecasting. For smaller datasets like SD, PreMixer surpasses the performance of baseline methods, except for DGCRN and \(\mathrm{D}^{2}\mathrm{STGNN}\). Additionally, PreMixer outperforms nearly all baselines in forecasting precision across predicted time steps on the GBA and GLA datasets. When considering the CA dataset, our model achieves exceptional performance, surpassing all others by a significant margin across every performance measure and time horizon. Thanks to its concise architecture, which eschews GNNs and sequential models, PreMixer delivers promising results for both effectiveness and efficiency across all datasets.

\subsection{Transfer Learning (RQ2)}
\begin{table*}[htbp]
\centering
\caption{Comparative evaluation results of transfer learning. The SD, GBA, and CA are selected as target datasets. The pre-trained model, initially trained on the source dataset, is transferred to the target dataset for traffic prediction. In each column, the bolded results represent traffic prediction conducted without transfer learning.}
\label{tab:transfer}
    \resizebox{\linewidth}{!}{%
        \begin{tabular}{cc|ccc|ccc|ccc|ccc}
        \toprule
             \multirow{2}{*}{ \textbf{Source} } & \multirow{2}{*}{ \textbf{Target} } & \multicolumn{3}{c}{\textbf{Horizon 3}} & \multicolumn{3}{c}{ \textbf{Horizon 6}} & \multicolumn{3}{c}{ \textbf{Horizon 12}} & \multicolumn{3}{c}{ \textbf{Average}} \\
            \cmidrule(r){3-14}
             & & MAE & RMSE & MAPE & MAE & RMSE & MAPE & MAE & RMSE & MAPE & MAE & RMSE & MAPE \\
             \hline \hline \rm{\textbf{SD}}
             &  \multirow{3}{*}{ \textbf{SD} } & \textbf{15.11} & \textbf{25.01} &  \textbf{9.78 \%}  & \textbf{17.88} & \textbf{29.81} &  \textbf{11.92 \%}  & \textbf{22.68} & \textbf{39.18} &  \textbf{15.75 \%}  & \textbf{18.02} & \textbf{30.22} &  \textbf{12.10 \%}   \\
             \rm{\textbf{GBA}} &   & 15.79 & 26.22 &  10.88 \%  & 18.87 & 31.68 & 13.86 \%  & 25.09 & 53.61 & 20.01\% & 19.35 & 35.63 &  14.51 \% \\
             \rm{\textbf{CA}} &  & 15.25 & 25.31 &  9.77 \%  & 18.17 & 30.47 &  12.09 \%  & 22.74 & 39.39 &  16.18 \%  & 18.22 & 30.86 &  12.39 \% \\
             
             \hline \hline \rm{\textbf{SD}} &  \multirow{3}{*}{ \textbf{GBA} }    & 17.40 & 28.95 &  13.44 \%  & 20.38 & 33.28 &  16.11 \%  & 24.43 & 39.33 &  20.00 \%  & 20.18 & 32.94 &  15.93 \%\\
             \rm{\textbf{GBA}} &   & \textbf{17.35} & \textbf{28.88} &  \textbf{13.03 \%}  & \textbf{20.21} & \textbf{33.15} &  \textbf{15.63 \%}  & \textbf{24.26} & \textbf{39.20} &  \textbf{19.87 \%}  & \textbf{20.06} & \textbf{32.85} &  \textbf{15.66 \%} \\
             \rm{\textbf{CA}} &  & 17.48 & 29.04 &  13.15 \%  & 20.32 & 33.19 &  15.75 \%  & 24.44 & 39.33 &  19.69 \%  & 20.19 & 32.95 &  15.66 \%\\

             \hline \hline  \rm{\textbf{SD}} &  \multirow{3}{*}{ \textbf{CA} }  & 15.84 & 26.36 &  11.44 \%  & 18.83 & 31.36 &  13.85 \%  & 23.14 & 38.62 &  17.85 \%  & 18.74 & 31.20 &  13.91\% \\
             \rm{\textbf{GBA}}&    & 15.69 & 26.18 &  11.39 \%  & 18.60 & 31.11 &  13.81 \%  & 22.72 & 38.31 &  17.35 \%  & 18.49 & 30.94 &  13.74 \%\\
             \rm{\textbf{CA}}&   & \textbf{15.56} & \textbf{25.97} &  \textbf{11.31 \%}  & \textbf{18.53} & \textbf{30.69} &  \textbf{13.57 \%}  & \textbf{22.59} & \textbf{37.56} &  \textbf{17.26 \%}  & \textbf{18.42} & \textbf{30.54} &  \textbf{13.60 \%} \\
        \bottomrule
        \end{tabular}
    }
\end{table*}
We conducted transfer learning experiments on three sub-datasets: SD, GBA, and CA. We trained the corresponding pre-trained models on the source dataset and then transferred them to the target training sets for prediction. Table \ref{tab:transfer} displays the test results of transfer learning and compares them to those of non-transfer learning in terms of MAE, RMSE, and MAPE, as well as the mean performance across 12 predicted time step. For the smaller SD target dataset, the transfer effect from the CA dataset is superior. This improvement likely arises from the CA dataset's larger size, which encompasses most features of the target datasets, thus enabling the pre-training process to capture the majority of traffic flow changes in those regions. For the GBA target dataset, the performance of pre-trained models using the small SD dataset and the large CA dataset as sources is similar, and closely aligns with the results observed without transfer. Transfers from smaller to larger regions, such as from GBA to CA, yield better outcomes. The GBA dataset, covering a larger geographical area and containing more nodes than the SD dataset, supports the development of a more robust pre-training model. However, cross-regional transfer learning, such as transfers between SD and GBA, tends to show poorer results, likely due to significant differences in regional data distributions.

\subsection{Ablation Study (RQ3)}
To determine the source of performance improvement, we conduct an ablation study on the PreMixer. These experiments entail designing and assessing several variants of PreMixer to check the impact of each modification on the overall performance of our model:

\begin{enumerate}
\item w/o Pre-training: This variant eliminates the entire pre-training module, allowing the model to make predictions without additional auxiliary information.
\item w/o CL: This variant retains the pre-training module but removes the contrastive learning in the pre-training stage.
\item w/o Context: This variant performs mixing operations only by the TemporalMixer and SpatialMixer, ignoring the spatio-temporal positional encoding and learnable node embedding.
\item w/o STPE: This variant uses typical sinusoidal positional encoding to replace Spatio-Temporal Positional Encoding.
\end{enumerate}

Table \ref{tab:Ablation experiments} presents the comparative experimental results of ablation study on the SD and GBA datasets. The PreMixer, equipped with auxiliary information, significantly outperforms its ablated versions, highlighting the crucial role of patch-level representation generated by PIEncoder. Furthermore, contrastive learning effectively enhances patch-wise time series representation by incorporating adjacent time series information. Additionally, the exploration of extending the MLP-Mixer paradigm to large-scale spatiotemporal urban data shows considerable potential. We combine spatio-temporal positional encoding with PreMixer to capture spatiotemporal heterogeneity, and the results demonstrate that this integration significantly improves the model’s prediction performance. The contextual information also proves helpful in addressing the inherent indistinguishability of samples. Overall, these findings underscore the value of our proposed MLP-based pre-training strategy and MLP-Mixer variants for traffic forecasting.

\begin{table}[htbp]
\centering
\caption{Ablation experiments. The results displayed in bold on the first row represent the performance of the model without any modifications or deletions.}
\label{tab:Ablation experiments}
\resizebox{\linewidth}{!}
{\begin{tabular}{lcccccc}
\toprule
Data &  \multicolumn{3}{c}{ SD} & \multicolumn{3}{c}{ GBA} \\
Metric & MAE & RMSE & MAPE & MAE & RMSE & MAPE \\
\midrule
PreMixer   & \textbf{18.02} & \textbf{30.22} &  \textbf{12.10 \%}  & \textbf{20.06} & \textbf{32.85} &  \textbf{15.66 \%}\\
w/o Pre-training   & 18.39 & 31.82  & 12.61\%  &20.39  &33.32  &15.88\%\\
w/o CL  &18.20 &30.64  &12.35\%  &20.15  &32.97  &15.74\% \\
w/o Context &19.34  &32.73  &13.10\%  &20.96  &34.10  &16.37\% \\
w/o STPE  &18.96  &31.63  &12.92\%  &20.70  &33.64  &16.06\%\\
\bottomrule
\end{tabular}}
\end{table}

\begin{table*}[htbp]
\centering
\caption{Efficiency comparisons. BS: batch size set during training. Train: training time (in seconds) per epoch. Infer: inference time (in seconds) on the validation set. Total: total training time (in hours). Note that the total training time is also dependent on the number of epochs completed.}
\label{tab:result_time}
    \resizebox{\linewidth}{!}{
        \begin{tabular}{c|cccc|cccc|cccc|cccc}
            \toprule \multirow{2}{*}{ Method } & \multicolumn{4}{c}{SD} & \multicolumn{4}{c}{GBA} & \multicolumn{4}{c}{GLA} & \multicolumn{4}{c}{ CA } \\
            \cmidrule(r){2-17}
             &  BS & Train & Infer & Total & BS & Train & Infer & Total &BS & Train & Infer & Total &BS & Train & Infer & Total \\
             \hline LSTM & 64 & 22 & 6 & 1 & 64 & 113 & 18 & 4 & 64 & 186 & 31 & 6 & 32 & 408 & 61 & 13 \\
             \hline ASTGCN & 64 & 135 & 21 & 4 & 40 & 1106 & 146 & 35 & 16 & 2985 & 382 & 94 & - & - & - & - \\
             \hline DCRNN & 64 & 865 & 150 & 28 & 64 & 1819 & 324 & 60 & 32 & 2588 & 453 & 84 & 16 & 4995 & 883 & 163 \\
             \hline AGCRN & 64 & 94 & 16 & 3 & 64 & 560 & 92 & 18 & 32 & 1447 & 251 & 41 & - & - & - & - \\
             \hline STGCN & 64 & 57 & 18 & 2 & 64 & 200 & 69 & 7 & 64 & 344 & 111 & 13 & 64 & 627 & 187 & 18 \\
             \hline GWNET & 64 & 96 & 14 & 3 & 64 & 484 & 67 & 11 & 64 & 1029 & 154 & 33 & 32 & 4096 & 549 & 129 \\
             \hline STGODE & 64 & 188 & 26 & 6 & 48 & 703 & 101 & 22 & 28 & 1305 & 190 & 25 & 12 & 4218 & 655 & 126 \\
             \hline DSTAGNN & 64 & 258 & 29 & 6 & 24 & 1983 & 175 & 41 & 10 & 5209 & 462 & 98 & - & - & - & - \\
             \hline DGCRN & 64 & 540 & 76 & 17 & 12 & 4191 & 606 & 56 & 5 & 10297 & 1928 & 207 & - & - & - & - \\
             \hline  $\rm{{D}^{2} STGNN}$ & 36 & 564 & 68 & 18 & 4 & 5748 & 758 & 145 & 4 & 5602 & 762 & 141 & - & - & - & - \\
             \hline PreMixer & \textbf{64} & \textbf{12} & \textbf{2} & \textbf{0.4} & \textbf{64} & \textbf{24} & \textbf{7} & \textbf{0.9} & \textbf{64} & \textbf{39} & \textbf{12} & \textbf{1.5} & \textbf{64} & \textbf{92} & \textbf{24} & \textbf{4}\\
            \bottomrule
        \end{tabular}
    }
\end{table*}

\subsection{Efficiency Comparisons (RQ4)}

\begin{figure}[htbp]
    \centering
    \includegraphics[width=0.5\textwidth]{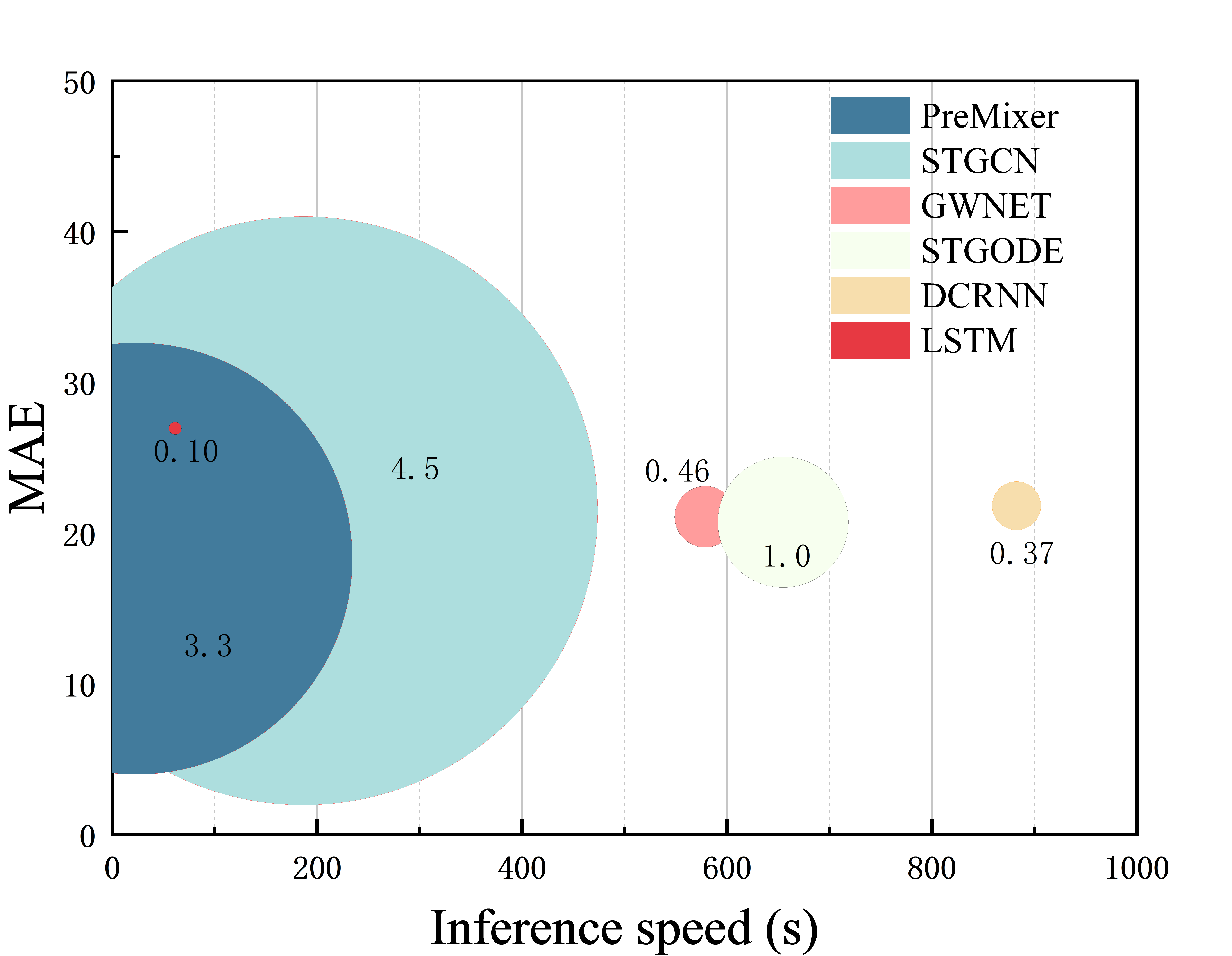}
    \caption{Comparison of PreMixer with baseline models across three key metrics: trainable parameters, inference speed, and average MAE in CA. The circle size in the figure corresponds to the trainable parameter number in each model.}
  \label{fig:compartive result}
\end{figure}

Table \ref{tab:result_time} presents a comparative efficiency analysis of the experimental times between our proposed model and several baseline models across four sub-datasets, detailing the training and inference time per epoch as well as the total training time. LSTM is faster than other baseline models due to its relatively simple technical architecture. Relying on the parallel computation of temporal convolution operations, STGCN and GWNET require comparatively less time and are scalable for large-scale datasets. Although DGCRN and \(\mathrm{D}^{2}\mathrm{STGNN}\) achieve good prediction results on all datasets, they are time-consuming and cannot be conducted on the CA dataset. For all baseline models, both training and inference times rapidly increase as the scale of the dataset expands, owing to their complex model architectures. Moreover, most baseline models cannot run on the CA dataset even utilize a small batch size, requiring significant memory storage. It is worth noting that PreMixer demonstrate significantly faster training and inference speed than all baseline models across all datasets, indicating its high efficiency and scalability. Fig.\ref{fig:compartive result} further emphasizes the prominent equilibrium between prediction accuracy and inference efficiency of our methodology. The size of the circles in the figure corresponds to the number of trainable parameters in each model. Despite its substantial number of parameters compared to simpler models, the PreMixer has achieved impressive prediction results on the large-scale CA dataset while maintaining reasonable inference times. This performance demonstrates the model's exceptional effectiveness and efficiency.

\subsection{Model Analysis (RQ5)}

\begin{figure}[htbp]
  \centering
  \includegraphics[width=0.5\textwidth]{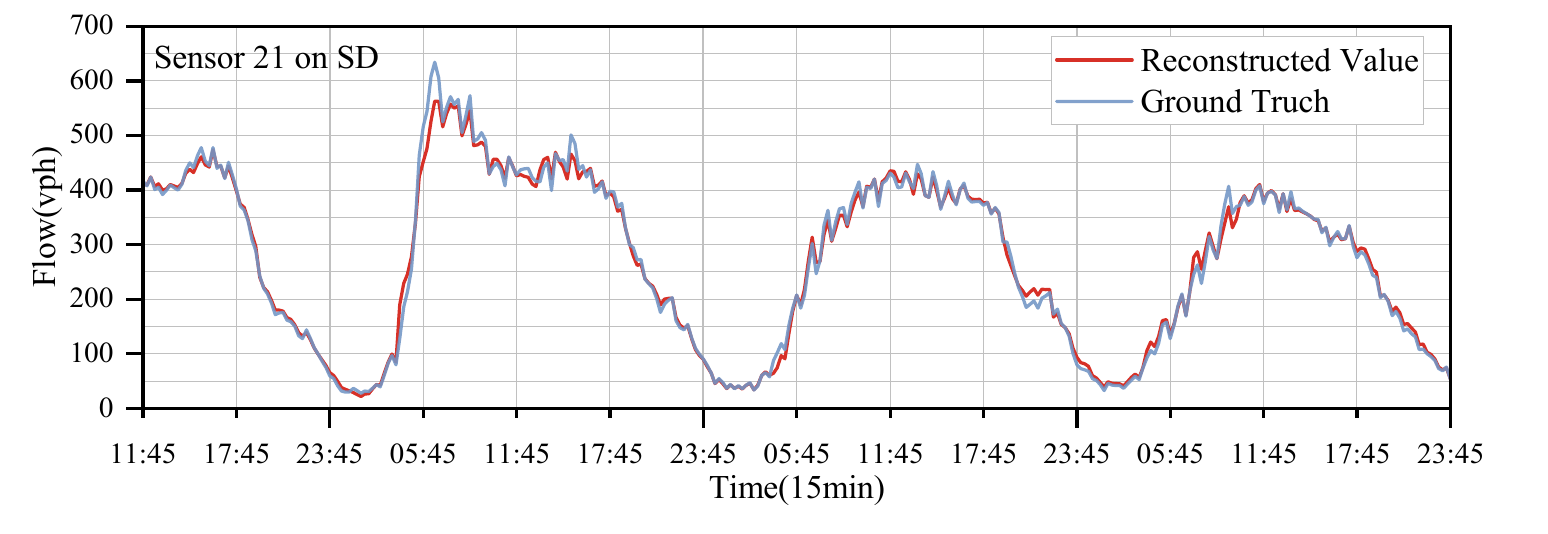}
  \caption{Reconstruction visualizations.}
  \label{fig:ron}
\end{figure}
\begin{figure*}[htbp]
     \centering
     \subfigure[SD dataset (Sensor ID = 1113181)]{\includegraphics[width=\textwidth]{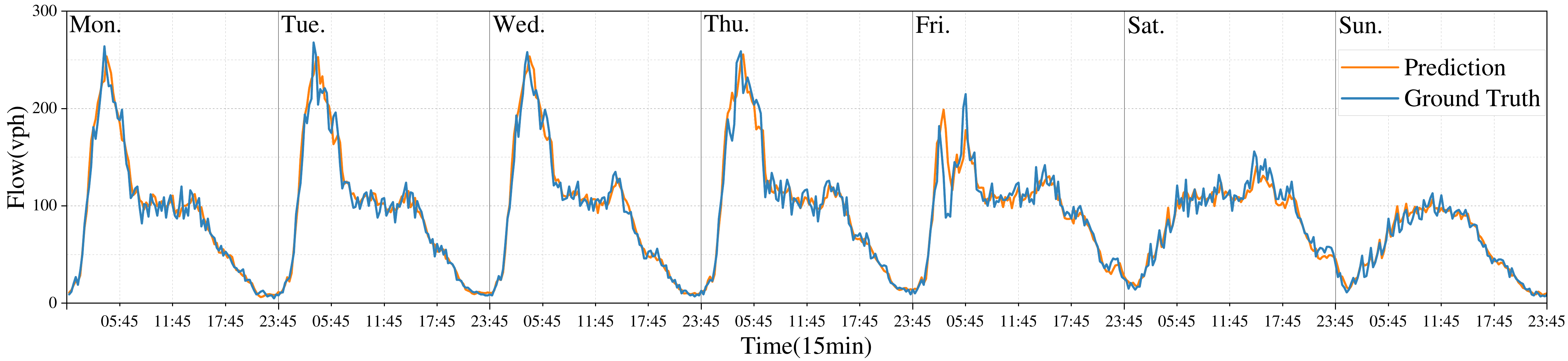}
	\label{SD}
	}
     \newline
     \subfigure[GBA dataset (Sensor ID = 402210)]{\includegraphics[width=\textwidth]{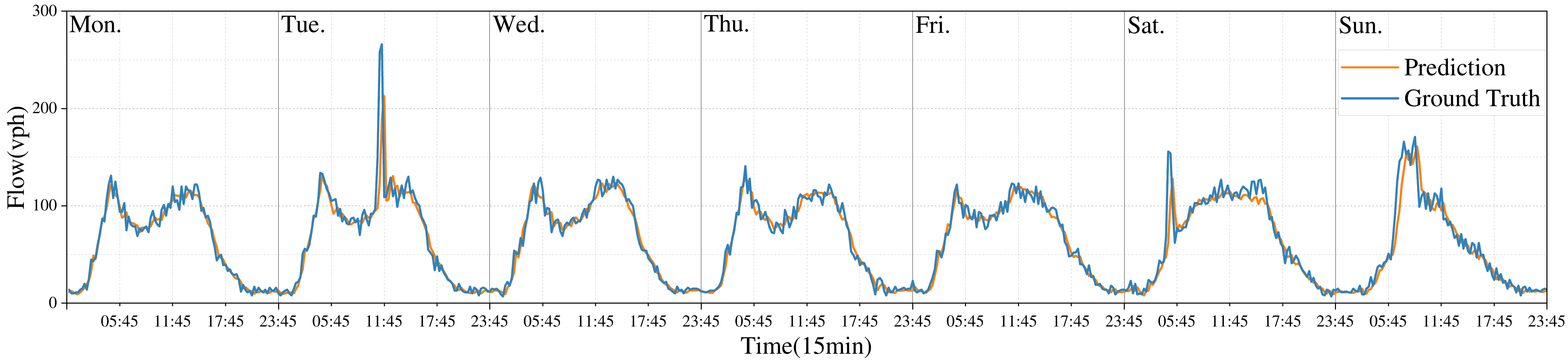}
	\label{GBA}
	}
     \newline
     \subfigure[CA dataset (Sensor ID = 315054)]{\includegraphics[width=\textwidth]{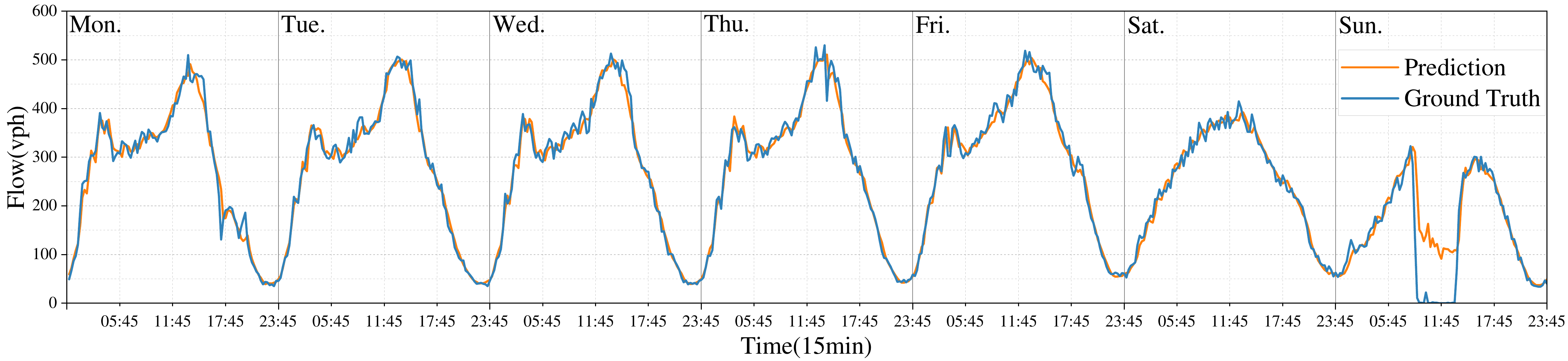}
	\label{CA}
	}
     \newline
    
    \caption{Comparison of the ground truth and the flow predicted by PreMixer tested on the LargeST dataset. The three sub-figures all show results from October 21, 2019, to October 27, 2019.}
    \label{fig:resultComparison}
\end{figure*}

To enhance understanding and evaluation of our model, we provide additional visualizations as shown below. Firstly, the visualization of the reconstruction outcomes by the PIEncoder on the dataset of SD is displayed in Figure \ref{fig:ron}. The PIEncoder can accurately reconstruct all patches even without considering information from surrounding time series patches. These findings confirm the robust capability of our model to extract contextual representations from traffic time series data with higher temporal coverage. However, in capturing the peak positions of traffic flow changes, the PIEncoder struggles to fully learn the underlying temporal patterns of traffic. Subsequently, we present visual comparisons of the predicted flow values and ground truth for the SD, GBA, and CA datasets to illustrate the predictive performance of the PreMixer, selecting time series that span one week. The time periods range from October 21, 2019, to October 27, 2019. The forecasting results are illustrated in Figure \ref{fig:resultComparison}, demonstrating that our model can reliably predict short-term traffic data across various temporal windows. Additionally, the model effectively identifies distinct temporal patterns during weekdays and weekends, and captures periodic traffic flow patterns while addressing local traffic fluctuations due to noise or other external factors. For instance, in the SD dataset, the model distinguishes the morning rush hour on Friday, which is significantly different from other days, demonstrating its ability to resist noise and make accurate predictions without overfitting. Similarly, in the largest CA dataset, the model's predictions for Sunday traffic confirm its robustness, although with some loss of accuracy.

\section{Conclusion}
\label{section6}
In this study, we propose PreMixer, an all-MLP traffic forecasting framework that seeks to offer simple yet effective methods for addressing critical challenges in large-scale traffic forecasting. The PreMixer provides a comprehensive framework designed for both pre-training and forecasting within large-scale traffic networks, showcasing outstanding efficiency, remarkable ability to capture contextual information, and scalability. In the pre-training phase, a novel patch reconstruction approach based on MLP is employed to efficiently decipher temporal patterns from extensive historical traffic data. For the forecasting phase, the downstream MLP-Mixer model benefits from the enhancements offered by pre-training model that independently embeds time series patches, thereby generating rich contextual patch-level representations. Additionally, we integrate spatio-temporal positional encoding and learnable node embedding to enhance the forecasting capabilities of MLP-Mixers for large-scale traffic prediction. Through comprehensive testing experiments, PreMixer has been demonstrated to surpass leading-edge methods, outperforming even simple LSTM models in large-scale traffic forecasting tasks.

Looking ahead, several promising avenues for future research are emerging: 
\begin{itemize}
  \item Forecasting within actual production environments presents unique challenges. While existing traffic prediction models have been extensively validated on public benchmarks, their effectiveness in actual real-world settings remains less explored. Factors such as meteorological conditions, community activities, and infrastructure projects significantly influence urban planning and traffic flow management. Overcoming these factors and implementing online learning technologies to adapt to dynamic environmental changes represents a promising direction for future research.
  \item Forecasting with external multimodal inputs can significantly enhance the precision and flexibility of traffic forecasting models. This approach is particularly effective in complex dynamic environments, leveraging multi-source data integration. Future research should focus on exploring dynamic multimodal fusion and cross-domain applications to further increase the flexibility and versatility of these models.
\end{itemize}

\bibliography{reference}

\begin{thebibliography}{10}
\providecommand{\url}[1]{#1}
\csname url@samestyle\endcsname
\providecommand{\newblock}{\relax}
\providecommand{\bibinfo}[2]{#2}
\providecommand{\BIBentrySTDinterwordspacing}{\spaceskip=0pt\relax}
\providecommand{\BIBentryALTinterwordstretchfactor}{4}
\providecommand{\BIBentryALTinterwordspacing}{\spaceskip=\fontdimen2\font plus
\BIBentryALTinterwordstretchfactor\fontdimen3\font minus \fontdimen4\font\relax}
\providecommand{\BIBforeignlanguage}[2]{{%
\expandafter\ifx\csname l@#1\endcsname\relax
\typeout{** WARNING: IEEEtran.bst: No hyphenation pattern has been}%
\typeout{** loaded for the language `#1'. Using the pattern for}%
\typeout{** the default language instead.}%
\else
\language=\csname l@#1\endcsname
\fi
#2}}
\providecommand{\BIBdecl}{\relax}
\BIBdecl

\bibitem{zheng2014urban}
Y.~Zheng, L.~Capra, O.~Wolfson, and H.~Yang, ``Urban computing: concepts, methodologies, and applications,'' \emph{ACM Transactions on Intelligent Systems and Technology (TIST)}, vol.~5, pp. 1--55, 2014.

\bibitem{polson2017deep}
N.~G. Polson and V.~O. Sokolov, ``Deep learning for short-term traffic flow prediction,'' \emph{Transportation Research Part C: Emerging Technologies}, vol.~79, pp. 1--17, 2017.

\bibitem{wu2004travel}
C.-H. Wu, J.-M. Ho, and D.-T. Lee, ``Travel-time prediction with support vector regression,'' \emph{IEEE transactions on intelligent transportation systems}, vol.~5, pp. 276--281, 2004.

\bibitem{cai2016spatiotemporal}
P.~Cai, Y.~Wang, G.~Lu, P.~Chen, C.~Ding, and J.~Sun, ``A spatiotemporal correlative k-nearest neighbor model for short-term traffic multistep forecasting,'' \emph{Transportation Research Part C: Emerging Technologies}, vol.~62, pp. 21--34, 2016.

\bibitem{ma2015long}
X.~Ma, Z.~Tao, Y.~Wang, H.~Yu, and Y.~Wang, ``Long short-term memory neural network for traffic speed prediction using remote microwave sensor data,'' \emph{Transportation Research Part C: Emerging Technologies}, vol.~54, pp. 187--197, 2015.

\bibitem{cui2020graph}
Z.~Cui, L.~Lin, Z.~Pu, and Y.~Wang, ``Graph markov network for traffic forecasting with missing data,'' \emph{Transportation Research Part C: Emerging Technologies}, vol. 117, p. 102671, 2020.

\bibitem{cui2020stacked}
Z.~Cui, R.~Ke, Z.~Pu, and Y.~Wang, ``Stacked bidirectional and unidirectional lstm recurrent neural network for forecasting network-wide traffic state with missing values,'' \emph{Transportation Research Part C: Emerging Technologies}, vol. 118, p. 102674, 2020.

\bibitem{WANG2024102293}
\BIBentryALTinterwordspacing
Z.~Wang, Y.~Wang, F.~Jia, F.~Zhang, N.~Klimenko, L.~Wang, Z.~He, Z.~Huang, and Y.~Liu, ``Spatiotemporal fusion transformer for large-scale traffic forecasting,'' \emph{Information Fusion}, vol. 107, p. 102293, 2024. [Online]. Available: \url{https://www.sciencedirect.com/science/article/pii/S156625352400071X}
\BIBentrySTDinterwordspacing

\bibitem{graves2012long}
A.~Graves and A.~Graves, ``Long short-term memory,'' \emph{Supervised sequence labelling with recurrent neural networks}, pp. 37--45, 2012.

\bibitem{bai2018empirical}
S.~Bai, J.~Z. Kolter, and V.~Koltun, ``An empirical evaluation of generic convolutional and recurrent networks for sequence modeling,'' \emph{arXiv preprint arXiv:1803.01271}, 2018.

\bibitem{vaswani2017attention}
A.~Vaswani, N.~Shazeer, N.~Parmar, J.~Uszkoreit, L.~Jones, A.~N. Gomez, {\L}.~Kaiser, and I.~Polosukhin, ``Attention is all you need,'' \emph{Advances in neural information processing systems}, vol.~30, 2017.

\bibitem{jin2023spatio}
G.~Jin, Y.~Liang, Y.~Fang, Z.~Shao, J.~Huang, J.~Zhang, and Y.~Zheng, ``Spatio-temporal graph neural networks for predictive learning in urban computing: A survey,'' \emph{IEEE Transactions on Knowledge and Data Engineering}, 2023.

\bibitem{liu2023we}
X.~Liu, Y.~Liang, C.~Huang, H.~Hu, Y.~Cao, B.~Hooi, and R.~Zimmermann, ``Do we really need graph neural networks for traffic forecasting?'' \emph{arXiv preprint arXiv:2301.12603}, 2023.

\bibitem{jiang2022graph}
W.~Jiang and J.~Luo, ``Graph neural network for traffic forecasting: A survey,'' \emph{Expert systems with applications}, vol. 207, p. 117921, 2022.

\bibitem{10.5555/3367243.3367303}
Z.~Wu, S.~Pan, G.~Long, J.~Jiang, and C.~Zhang, ``Graph wavenet for deep spatial-temporal graph modeling,'' in \emph{Proceedings of the 28th International Joint Conference on Artificial Intelligence}, ser. IJCAI'19.\hskip 1em plus 0.5em minus 0.4em\relax AAAI Press, 2019, p. 1907–1913.

\bibitem{10032279}
Y.~Wang, C.~Jing, W.~Huang, S.~Jin, and X.~Lv, ``Adaptive spatiotemporal inceptionnet for traffic flow forecasting,'' \emph{IEEE Transactions on Intelligent Transportation Systems}, vol.~24, pp. 3882--3907, 2023.

\bibitem{pan2020spatio}
Z.~Pan, W.~Zhang, Y.~Liang, W.~Zhang, Y.~Yu, J.~Zhang, and Y.~Zheng, ``Spatio-temporal meta learning for urban traffic prediction,'' \emph{IEEE Transactions on Knowledge and Data Engineering}, vol.~34, pp. 1462--1476, 2020.

\bibitem{ji2023spatio}
J.~Ji, J.~Wang, C.~Huang, J.~Wu, B.~Xu, Z.~Wu, J.~Zhang, and Y.~Zheng, ``Spatio-temporal self-supervised learning for traffic flow prediction,'' in \emph{Proceedings of the AAAI conference on artificial intelligence}, vol.~37, 2023, pp. 4356--4364.

\bibitem{jiang2023pdformer}
J.~Jiang, C.~Han, W.~X. Zhao, and J.~Wang, ``Pdformer: Propagation delay-aware dynamic long-range transformer for traffic flow prediction,'' in \emph{Proceedings of the AAAI conference on artificial intelligence}, vol.~37, 2023, pp. 4365--4373.

\bibitem{xu2020spatial}
M.~Xu, W.~Dai, C.~Liu, X.~Gao, W.~Lin, G.-J. Qi, and H.~Xiong, ``Spatial-temporal transformer networks for traffic flow forecasting,'' \emph{arXiv preprint arXiv:2001.02908}, 2020.

\bibitem{mallick2020graph}
T.~Mallick, P.~Balaprakash, E.~Rask, and J.~Macfarlane, ``Graph-partitioning-based diffusion convolutional recurrent neural network for large-scale traffic forecasting,'' \emph{Transportation Research Record}, vol. 2674, pp. 473--488, 2020.

\bibitem{cini2024taming}
A.~Cini, I.~Marisca, D.~Zambon, and C.~Alippi, ``Taming local effects in graph-based spatiotemporal forecasting,'' \emph{Advances in Neural Information Processing Systems}, vol.~36, 2024.

\bibitem{clark2016changes}
B.~Clark, K.~Chatterjee, and S.~Melia, ``Changes to commute mode: The role of life events, spatial context and environmental attitude,'' \emph{Transportation Research Part A: Policy and Practice}, vol.~89, pp. 89--105, 2016.

\bibitem{liu2024largest}
X.~Liu, Y.~Xia, Y.~Liang, J.~Hu, Y.~Wang, L.~Bai, C.~Huang, Z.~Liu, B.~Hooi, and R.~Zimmermann, ``Largest: A benchmark dataset for large-scale traffic forecasting,'' \emph{Advances in Neural Information Processing Systems}, vol.~36, 2024.

\bibitem{zhou2020graph}
J.~Zhou, G.~Cui, S.~Hu, Z.~Zhang, C.~Yang, Z.~Liu, L.~Wang, C.~Li, and M.~Sun, ``Graph neural networks: A review of methods and applications,'' \emph{AI open}, vol.~1, pp. 57--81, 2020.

\bibitem{wang2024spatiotemporal}
Z.~Wang, Y.~Wang, F.~Jia, F.~Zhang, N.~Klimenko, L.~Wang, Z.~He, Z.~Huang, and Y.~Liu, ``Spatiotemporal fusion transformer for large-scale traffic forecasting,'' \emph{Information Fusion}, vol. 107, p. 102293, 2024.

\bibitem{10.1145/3534678.3539396}
\BIBentryALTinterwordspacing
Z.~Shao, Z.~Zhang, F.~Wang, and Y.~Xu, ``Pre-training enhanced spatial-temporal graph neural network for multivariate time series forecasting,'' in \emph{Proceedings of the 28th ACM SIGKDD Conference on Knowledge Discovery and Data Mining}, ser. KDD '22.\hskip 1em plus 0.5em minus 0.4em\relax New York, NY, USA: Association for Computing Machinery, 2022, p. 1567–1577. [Online]. Available: \url{https://doi.org/10.1145/3534678.3539396}
\BIBentrySTDinterwordspacing

\bibitem{shao2022spatial}
Z.~Shao, Z.~Zhang, F.~Wang, W.~Wei, and Y.~Xu, ``Spatial-temporal identity: A simple yet effective baseline for multivariate time series forecasting,'' in \emph{Proceedings of the 31st ACM International Conference on Information \& Knowledge Management}, 2022, pp. 4454--4458.

\bibitem{he2022masked}
K.~He, X.~Chen, S.~Xie, Y.~Li, P.~Doll{\'a}r, and R.~Girshick, ``Masked autoencoders are scalable vision learners,'' in \emph{Proceedings of the IEEE/CVF conference on computer vision and pattern recognition}, 2022, pp. 16\,000--16\,009.

\bibitem{qiu2020pre}
X.~Qiu, T.~Sun, Y.~Xu, Y.~Shao, N.~Dai, and X.~Huang, ``Pre-trained models for natural language processing: A survey,'' \emph{Science China technological sciences}, vol.~63, pp. 1872--1897, 2020.

\bibitem{gao2023spatio}
H.~Gao, R.~Jiang, Z.~Dong, J.~Deng, and X.~Song, ``Spatio-temporal-decoupled masked pre-training for traffic forecasting,'' \emph{arXiv preprint arXiv:2312.00516}, 2023.

\bibitem{tolstikhin2021mlp}
I.~O. Tolstikhin, N.~Houlsby, A.~Kolesnikov, L.~Beyer, X.~Zhai, T.~Unterthiner, J.~Yung, A.~Steiner, D.~Keysers, J.~Uszkoreit \emph{et~al.}, ``Mlp-mixer: An all-mlp architecture for vision,'' \emph{Advances in neural information processing systems}, vol.~34, pp. 24\,261--24\,272, 2021.

\bibitem{fusco2023pnlp}
F.~Fusco, D.~Pascual, P.~Staar, and D.~Antognini, ``pnlp-mixer: an efficient all-mlp architecture for language,'' in \emph{Proceedings of the 61st Annual Meeting of the Association for Computational Linguistics (Volume 5: Industry Track)}, 2023, pp. 53--60.

\bibitem{chen2023tsmixer}
S.-A. Chen, C.-L. Li, N.~Yoder, S.~O. Arik, and T.~Pfister, ``Tsmixer: An all-mlp architecture for time series forecasting,'' \emph{arXiv preprint arXiv:2303.06053}, 2023.

\bibitem{yeh2024random}
C.-C.~M. Yeh, Y.~Fan, X.~Dai, V.~Lai, P.~O. Aboagye, J.~Wang, H.~Chen, Y.~Zheng, Z.~Zhuang, L.~Wang \emph{et~al.}, ``Random projection layers for multidimensional time sires forecasting,'' \emph{arXiv preprint arXiv:2402.10487}, 2024.

\bibitem{wang2023st}
Z.~Wang, Y.~Nie, P.~Sun, N.~H. Nguyen, J.~Mulvey, and H.~V. Poor, ``St-mlp: A cascaded spatio-temporal linear framework with channel-independence strategy for traffic forecasting,'' \emph{arXiv preprint arXiv:2308.07496}, 2023.

\bibitem{nie2023contextualizing}
T.~Nie, G.~Qin, L.~Sun, W.~Ma, Y.~Mei, and J.~Sun, ``Contextualizing mlp-mixers spatiotemporally for urban data forecast at scale,'' \emph{arXiv preprint arXiv:2307.01482}, 2023.

\bibitem{lan2022dstagnn}
S.~Lan, Y.~Ma, W.~Huang, W.~Wang, H.~Yang, and P.~Li, ``Dstagnn: Dynamic spatial-temporal aware graph neural network for traffic flow forecasting,'' in \emph{International conference on machine learning}.\hskip 1em plus 0.5em minus 0.4em\relax PMLR, 2022, pp. 11\,906--11\,917.

\bibitem{li2023dynamic}
F.~Li, J.~Feng, H.~Yan, G.~Jin, F.~Yang, F.~Sun, D.~Jin, and Y.~Li, ``Dynamic graph convolutional recurrent network for traffic prediction: Benchmark and solution,'' \emph{ACM Transactions on Knowledge Discovery from Data}, vol.~17, pp. 1--21, 2023.

\bibitem{guo2019attention}
S.~Guo, Y.~Lin, N.~Feng, C.~Song, and H.~Wan, ``Attention based spatial-temporal graph convolutional networks for traffic flow forecasting,'' in \emph{Proceedings of the AAAI conference on artificial intelligence}, vol.~33, 2019, pp. 922--929.

\bibitem{liu2022contrastive}
X.~Liu, Y.~Liang, C.~Huang, Y.~Zheng, B.~Hooi, and R.~Zimmermann, ``When do contrastive learning signals help spatio-temporal graph forecasting?'' in \emph{Proceedings of the 30th international conference on advances in geographic information systems}, 2022, pp. 1--12.

\bibitem{li2022mining}
R.~Li, T.~Zhong, X.~Jiang, G.~Trajcevski, J.~Wu, and F.~Zhou, ``Mining spatio-temporal relations via self-paced graph contrastive learning,'' in \emph{Proceedings of the 28th ACM SIGKDD Conference on Knowledge Discovery and Data Mining}, 2022, pp. 936--944.

\bibitem{dong2024simmtm}
J.~Dong, H.~Wu, H.~Zhang, L.~Zhang, J.~Wang, and M.~Long, ``Simmtm: A simple pre-training framework for masked time-series modeling,'' \emph{Advances in Neural Information Processing Systems}, vol.~36, 2024.

\bibitem{nie2022time}
Y.~Nie, N.~H. Nguyen, P.~Sinthong, and J.~Kalagnanam, ``A time series is worth 64 words: Long-term forecasting with transformers,'' \emph{arXiv preprint arXiv:2211.14730}, 2022.

\bibitem{wu2023interpretable}
H.~Wu, H.~Zhou, M.~Long, and J.~Wang, ``Interpretable weather forecasting for worldwide stations with a unified deep model,'' \emph{Nature Machine Intelligence}, vol.~5, no.~6, pp. 602--611, 2023.

\bibitem{wang2023pfnet}
C.~Wang, K.~Zuo, S.~Zhang, H.~Lei, P.~Hu, Z.~Shen, R.~Wang, and P.~Zhao, ``Pfnet: Large-scale traffic forecasting with progressive spatio-temporal fusion,'' \emph{IEEE Transactions on Intelligent Transportation Systems}, vol.~24, no.~12, pp. 14\,580--14\,597, 2023.

\bibitem{fang2021spatial}
Z.~Fang, Q.~Long, G.~Song, and K.~Xie, ``Spatial-temporal graph ode networks for traffic flow forecasting,'' in \emph{Proceedings of the 27th ACM SIGKDD conference on knowledge discovery \& data mining}, 2021, pp. 364--373.

\bibitem{zheng2020gman}
C.~Zheng, X.~Fan, C.~Wang, and J.~Qi, ``Gman: A graph multi-attention network for traffic prediction,'' in \emph{Proceedings of the AAAI conference on artificial intelligence}, vol.~34, 2020, pp. 1234--1241.

\bibitem{qin2023spatio}
Y.~Qin, H.~Luo, F.~Zhao, Y.~Fang, X.~Tao, and C.~Wang, ``Spatio-temporal hierarchical mlp network for traffic forecasting,'' \emph{Information Sciences}, vol. 632, pp. 543--554, 2023.

\bibitem{lee2023learning}
S.~Lee, T.~Park, and K.~Lee, ``Learning to embed time series patches independently,'' \emph{arXiv preprint arXiv:2312.16427}, 2023.

\bibitem{nonnenmacher2022utilizing}
M.~T. Nonnenmacher, L.~Oldenburg, I.~Steinwart, and D.~Reeb, ``Utilizing expert features for contrastive learning of time-series representations,'' in \emph{International Conference on Machine Learning}.\hskip 1em plus 0.5em minus 0.4em\relax PMLR, 2022, pp. 16\,969--16\,989.

\bibitem{lee2020mix}
K.~Lee, Y.~Zhu, K.~Sohn, C.-L. Li, J.~Shin, and H.~Lee, ``i-mix: A domain-agnostic strategy for contrastive representation learning,'' \emph{arXiv preprint arXiv:2010.08887}, 2020.

\bibitem{wang2021translating}
Z.~Wang and J.-C. Liu, ``Translating math formula images to latex sequences using deep neural networks with sequence-level training,'' \emph{International Journal on Document Analysis and Recognition (IJDAR)}, vol.~24, pp. 63--75, 2021.

\bibitem{chen2001freeway}
C.~Chen, K.~Petty, A.~Skabardonis, P.~Varaiya, and Z.~Jia, ``Freeway performance measurement system: mining loop detector data,'' \emph{Transportation research record}, vol. 1748, pp. 96--102, 2001.

\bibitem{liang2021revisiting}
Y.~Liang, K.~Ouyang, Y.~Wang, Y.~Liu, J.~Zhang, Y.~Zheng, and D.~S. Rosenblum, ``Revisiting convolutional neural networks for citywide crowd flow analytics,'' in \emph{Machine Learning and Knowledge Discovery in Databases: European Conference, ECML PKDD 2020, Ghent, Belgium, September 14--18, 2020, Proceedings, Part I}.\hskip 1em plus 0.5em minus 0.4em\relax Springer, 2021, pp. 578--594.

\bibitem{li2017diffusion}
Y.~Li, R.~Yu, C.~Shahabi, and Y.~Liu, ``Diffusion convolutional recurrent neural network: Data-driven traffic forecasting,'' \emph{arXiv preprint arXiv:1707.01926}, 2017.

\bibitem{bai2020adaptive}
L.~Bai, L.~Yao, C.~Li, X.~Wang, and C.~Wang, ``Adaptive graph convolutional recurrent network for traffic forecasting,'' \emph{Advances in neural information processing systems}, vol.~33, pp. 17\,804--17\,815, 2020.

\bibitem{10.5555/3304222.3304273}
B.~Yu, H.~Yin, and Z.~Zhu, ``Spatio-temporal graph convolutional networks: a deep learning framework for traffic forecasting,'' in \emph{Proceedings of the 27th International Joint Conference on Artificial Intelligence}, ser. IJCAI'18.\hskip 1em plus 0.5em minus 0.4em\relax AAAI Press, 2018, p. 3634–3640.

\bibitem{shao2022decoupled}
Z.~Shao, Z.~Zhang, W.~Wei, F.~Wang, Y.~Xu, X.~Cao, and C.~S. Jensen, ``Decoupled dynamic spatial-temporal graph neural network for traffic forecasting,'' \emph{Proceedings of the VLDB Endowment}, vol.~15, pp. 2733--2746, 2022.

\end{thebibliography}
\bibliographystyle{IEEEtran}

\begin{IEEEbiography}
[{\includegraphics[width=1in,height=1.25in,clip]{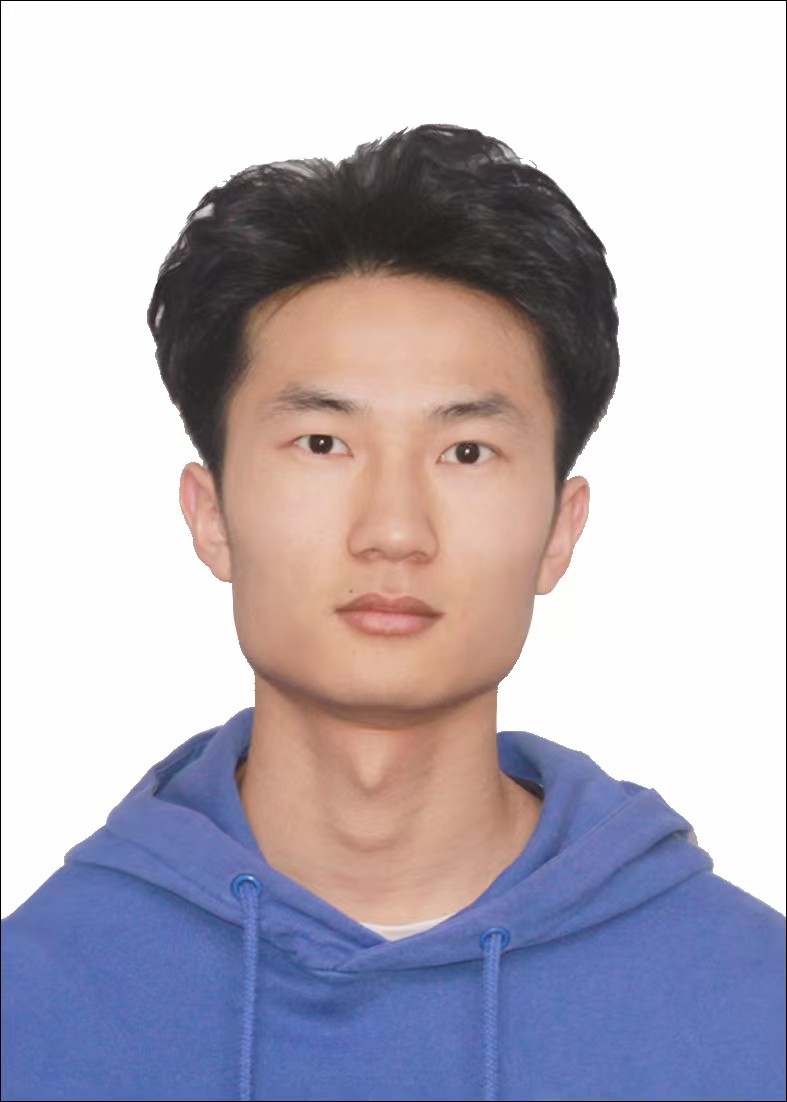}}]{Tongtong Zhang}
received the B.S. degree in automation from North China Electric Power University in 2021. He is currently pursuing a master’s degree in transportation at Beihang University. His main research interests include traffic prediction and deep learning.
\end{IEEEbiography}

\begin{IEEEbiography}
[{\includegraphics[width=1in,height=1.25in,clip]{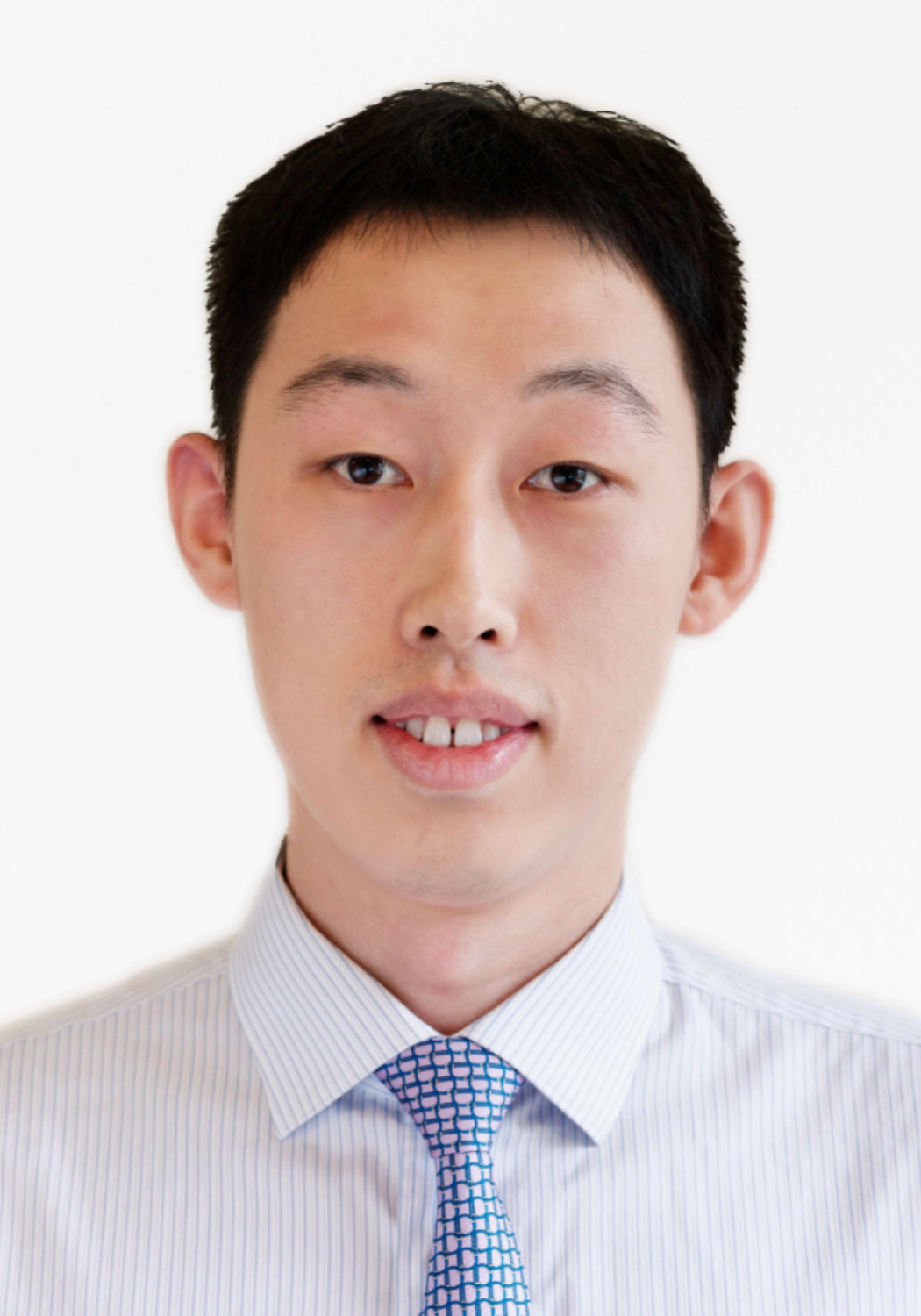}}]{Zhiyong Cui}
(Member, IEEE) received the B.S. degree in software engineering from Beihang University, Beijing, China, in 2012, the M.S. degree in software engineering and microelectronics from Peking University, Beijing, in 2015, and the Ph.D. degree in civil engineering from the University of Washington, Seattle, WA, USA, in 2021. He is currently a Professor in the School of Transportation Science and Engineering at Beihang University. His primary research interests include urban computing, traffic forecasting, and connected vehicles.
\end{IEEEbiography}

\begin{IEEEbiography}
[{\includegraphics[width=1in,height=1.25in,clip]{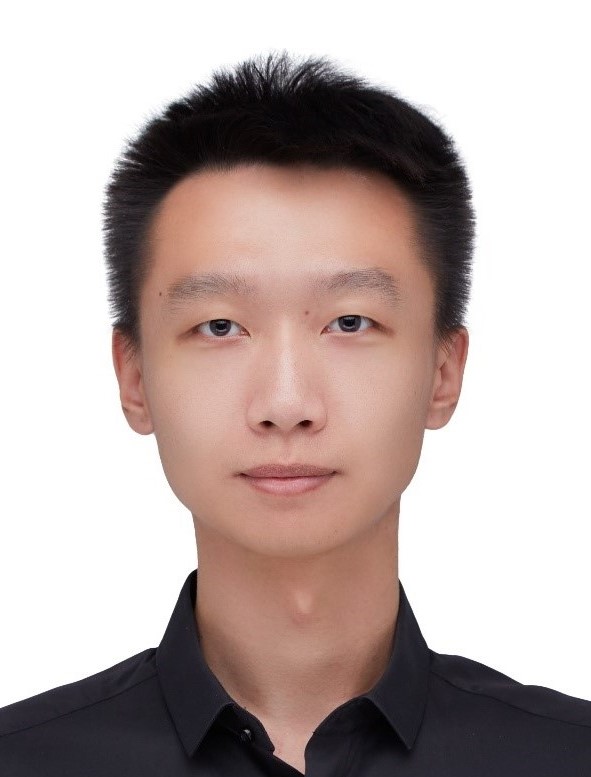}}]{Bingzhang Wang}
received a M.S. degree in Transportation Engineering from University of Washington in 2024, and double B.S. degrees in Mechanical Engineering from Shanghai Jiao Tong University in 2020 and Software Engineering from Peking University in 2022. His research direction is AI/ML in Transportation, with a focus on Large Language Models and Computer Vision techniques for traffic perception, analytics, and forecasting.
\end{IEEEbiography}

\begin{IEEEbiography}
[{\includegraphics[width=1in,height=1.25in,clip]{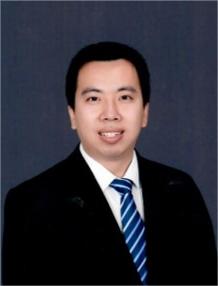}}]{Yilong Ren}
(Member, IEEE) received the B.S. and Ph.D. degrees from Beihang University in 2010 and 2017, respectively. He is currently an Associate Professor with the Research Institute for Frontier Science, Beihang University. His research interests include vehicular communications, vehicular crowd sensing, traffic big data, intelligent vehicle infrastructure cooperative systems, and intelligent vehicle intelligent testing.
\end{IEEEbiography}

\begin{IEEEbiography}
[{\includegraphics[width=1in,height=1.25in,clip]{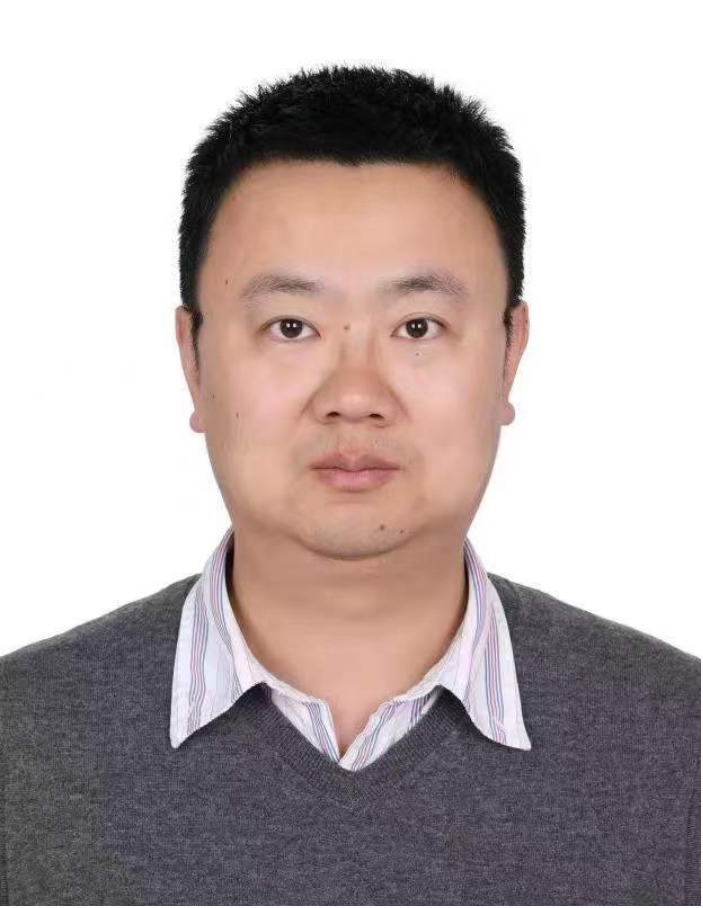}}]{Haiyang Yu}
(Member, IEEE) received the Ph.D. degree in traffic environment and safety technology from Jilin University, China, in 2012. He is currently a University Professor with the School of Transportation Science and Engineering, Beihang University, China. His research interests include traffic big data, traffic control, intelligent vehicle infrastructure cooperative systems, and intelligent vehicle intelligent testing.
\end{IEEEbiography}

\begin{IEEEbiography}
[{\includegraphics[width=1in,height=1.25in,clip]{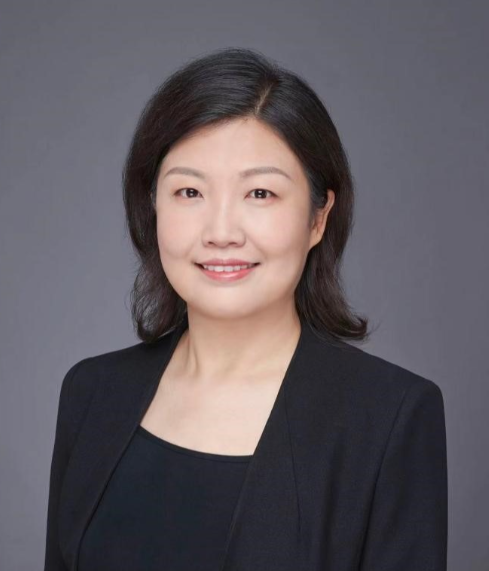}}]{Pan Deng}
received the B.S. degree in software engineering from the Harbin Institute of Technology, Harbin, China, in 2004, and the D.S. degree in computer science and technology from the Beijing University of Aeronautics and Astronautics, Beijing, China, in 2011. She is currently an Associate Research Fellow with the Laboratory of Parallel Software and Computational Science, Institute of Software, Chinese Academy of sciences, Beijing. She has directed over 19 science projects as a Project Leader, and has authored or co-authored over 25 scientific papers, four Chinese patents, and 16 software copyrights. Her research interests include grid computing, large-scale device collaboration, and formal methods.
\end{IEEEbiography}

\begin{IEEEbiography}
[{\includegraphics[width=1in,height=1.25in,clip]{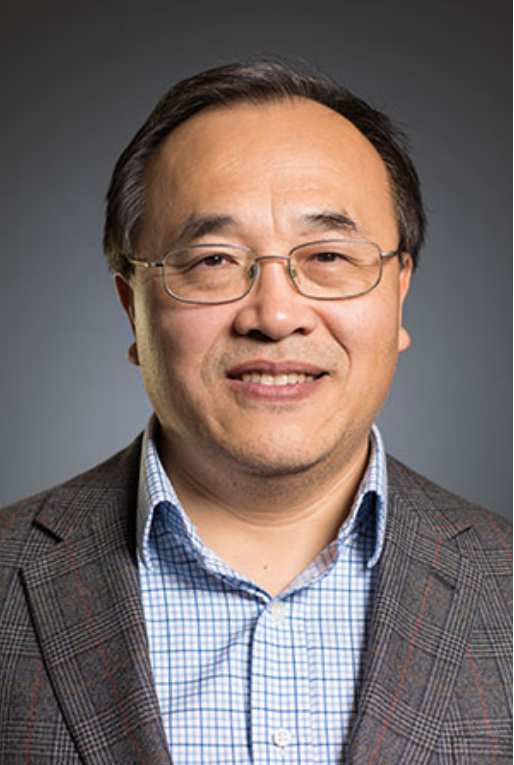}}]{Yinhai Wang}
received the master’s degree in computer science from the University of Washington (UW) and the Ph.D. degree in transportation engineering from The University of Tokyo in 1998. He is currently a professor in transportation engineering and the Founding Director of the Smart Transportation Applications and Research Laboratory (STAR Lab), UW. He also serves as the Director of the Pacific Northwest Transportation Consortium (PacTrans), U.S. Department of Transportation, University Transportation Center for Federal Region 10.
\end{IEEEbiography}

\vfill

\end{document}